\documentclass[5p,twocolumn]{elsarticle}
\usepackage[english]{babel}  
\usepackage{multirow}
\usepackage{comment}
\usepackage{subfig}
\usepackage{amsmath,amssymb,amsfonts}
\usepackage{graphicx}
\usepackage{pgfplots}
\usepackage{filecontents}
\usepackage{pgf-pie}
\usepackage{subcaption}
\usepackage{natbib}
\usetikzlibrary{backgrounds}
\usepackage{epstopdf}
\pgfplotsset{compat=1.18}

\begin{document}
\begin{frontmatter}

\title{PNN: A Novel Progressive Neural Network for Fault Classification in Rotating Machinery under Small Dataset Constraint}

\author{Praveen Chopra, Himanshu Kumar, Sandeep Yadav} 

\begin{abstract}
Fault detection in rotating machinery is a complex task, particularly in small and heterogeneous dataset scenarios. Variability in sensor placement, machinery configurations, and structural differences further increase the complexity of the problem. Conventional deep learning approaches often demand large, homogeneous datasets, limiting their applicability in data-scarce industrial environments. While transfer learning and few-shot learning have shown potential, however, they are often constrained by the need for extensive fault datasets. This research introduces a unified framework leveraging a novel progressive neural network (PNN) architecture designed to address these challenges. The PNN sequentially estimates the fixed-size refined features of the higher order with the help of all previously estimated features and appends them to the feature set. This fixed-size feature output at each layer controls the complexity of the PNN and makes it suitable for effective learning from small datasets. The framework's effectiveness is validated on eight datasets, including six open-source datasets, one in-house fault simulator, and one real-world industrial dataset. The PNN achieves state-of-the-art performance in fault detection across varying dataset sizes and machinery types, highlighting superior generalization and classification capabilities. 
\end{abstract}

\begin{keyword}
Small-size dataset, Fault detection, Open source dataset, Rotating machinery, Fault, Progressive Neural Network, Deep neural network, Implemented AI, Application of AI
\end{keyword}

\end{frontmatter}

\section{Introduction}\label{sec:Introduction}
\nocite{*}

Faults in rotating machinery often manifest as distinctive patterns in vibration and acoustic signals \cite{c1, c35}. A substantial body of work on fault detection and classification relies on vibration data acquired through strategically placed sensors \cite{c2, c3, c8, c9}. However, the effectiveness of fault detection can be significantly influenced by factors such as sensor placement, machinery configuration, and mechanical structure; making it challenging to generalize across different setups.

\par The fault-related features in vibration or acoustic data often exhibit intricate dependencies on operating frequencies and their harmonics \cite{c1, c35}. Traditional methods, such as wavelet-based feature extraction \cite{c24, c25, c26} and FFT spectrum-based statistical correlation \cite{c23}, have shown effectiveness, but these methods are typically tailored to specific datasets or machinery configurations. Variations in operating speeds, sampling rates, recording durations, and machinery types further complicate the generalization of these models. This is particularly problematic for traditional deep learning (DL) models, such as AEC (Auto Encoder) \cite{c41}, CNN (Convolutional Neural Network) \cite{c42}, LeNet \cite{c42}, AlexNet \cite{c45}, ResNet18 \cite{c38}, and LSTM (Long Short-Term Memory) \cite{c43}, which require large datasets for effective training. A comparative analysis in \cite{c3} using a standard 75–25\% train-test split on six open-source datasets highlights the limitations of traditional DL models. The advanced DL approaches based on these traditional DL models \cite{c3, c29, c30, c31} demonstrate high accuracy but are often dataset-specific and require large amounts of training datasets. 

\par Industrial scenarios present additional challenges due to limited and heterogeneous datasets. High-quality recordings for certain fault classes are often scarce, primarily due to sensor-related issues or setup limitations, resulting in small and uneven datasets.  While transfer learning offers improvements in small-data scenarios \cite{c47}, it often necessitates large source datasets and domain-specific knowledge for synthetic data generation, making it computationally intensive.

\par Few-shot learning (FSL) methods have emerged as promising alternatives for addressing small-data challenges. Techniques like Siamese CNNs \cite{c7} and denoising autoencoders (DAEs) with self-attention mechanisms \cite{c46} have demonstrated efficacy in fault classification tasks. For example, FSL applied to the CWRU dataset achieves high performance using a Siamese CNN-based network \cite{c7}. Other methods, such as multi-wavelet deep autoencoders \cite{c47} and MAMF-HGCN for aerospace applications \cite{c53}, reduce dependence on large datasets but still require substantial data for training robust feature extractors. Similarly, TabPFN \cite{c48}, a transformer-based model trained on a vast corpus of synthetic data and fine-tuned on target datasets, achieves high performance on small-size datasets. However, TabPFN's dependence on extensive synthetic data resources to fine-tune smaller datasets.

\par The recent DL-based advance model, the AMCMENet (Adaptive Multiscale Convolution Manifold Embedding Networks) \cite{c55}, a 12-layer large size DNN, with a large volume of training data achieves high accuracy on open source CWRU dataset. Similarly,  the NCVAE-AFL (Normalized Conditional Variational Auto-Encoder with Adaptive Focal Loss) model \cite{c52}, optimized for class-imbalanced datasets. This model achieves high accuracy at small-size training-testing ratios of $20-80\%$. However, NCVAE-AFL, being a large encoder-decoder VAE-based DNN, requires significantly higher training time due to its large size.

\par Recent DL-based advancements, including  UFADPC \cite{c20}, CORAL \cite{c9}, TCNN \cite{c8}, LSISMM \cite{c50}, AIICNN \cite{c51},  DGGCAE \cite{c54}, and UDTL \cite{c2}, have improved fault classification performance. All the above methods are basically deep neural networks (DNNs) with a large number of parameters. However, these DNNs remain heavily reliant on extensive training data and dataset-specific configurations, limiting their applicability in real-world industrial settings where data acquisition is resource-constrained.

\par While recent techniques have shown success in fault detection and classification, challenges remain when dealing with small-size datasets:
\begin{itemize}
    \item Most methods rely on large deep neural networks (DNNs), which require substantial labeled training data, either from rotary machines or synthetic sources.
    \item Transfer learning depends on a source domain with a large training dataset to effectively train the model.
    \item Even in few-shot learning, a significant amount of data is needed to train the feature extractor.
\end{itemize}

This paper addresses these limitations of these DNNs by proposing a generalized framework for fault detection and classification in small-data scenarios. The proposed model is a DNN with a very small number of parameters and requires a small size of the training dataset. The framework is evaluated on eight datasets, including six widely-used open-source datasets \cite{c2, c3}, an in-house machinery fault simulator dataset, and a dataset from a real industrial IC-Engine setup. The framework aligns with the Industry 4.0 paradigm by emphasizing compact model design for deployment on resource-constrained embedded systems without compromising accuracy.

The key contributions of this paper are:
\begin{enumerate}
    \item A novel Progressive Neural Network (PNN) architecture for fault classification for rotatory machines. The PNN sequentially estimates the fixed-size refined features of the higher order with the help of all previously estimated features and appends them to the feature set. This fixed-size feature output at each layer controls the complexity of the PNN and makes it suitable for effective learning from small datasets. Refined features with each layer ensure consistent performance across all evaluated eight datasets. 
    \item The paper devises a \textit{Dataset Standardization Technique} that first transforms the input time domain signal into the frequency domain counterpart using the Fast Fourier Transform (FFT). This output is further standardized by quantizing it to a fixed length using the maximum value among the bins. This not only reduces the size of the input but also keeps intact all the dominant features of the input fault signal. decouples the fault features, which are related to operating frequencies and their harmonics \cite{c1, c35}. Again, the dominant features in fixed bin size reduce the complexity of the learning due to ordered features. This is efficient for small-size datasets, as evidenced by the results. 
 \end{enumerate}
\par Following are the major innovative aspects of the PNN architecture:
\begin{enumerate}
    \item Sequential Feature Refinement: PNN sequentially estimates fixed-size refined features at each layer, incorporating information from all previously estimated features. This controlled feature growth enhances learning efficiency, particularly on small datasets.
    \item Minimal number of parameters, reducing computational complexity with mitigating overfitting and improving generalization on limited data..
\item Vanishing Gradient Mitigation: The sequential architecture inherently addresses the vanishing gradient problem, enabling stable training even with deeper networks.
    \item Robust feature extraction through progressive feature reuse, enhancing its suitability for small datasets.
\end{enumerate}

The organization of this article is as follows: Section 2 presents the problem formulation. The proposed methodology, with the description of the data preprocessing and the architecture of PNN, is given in section 3. Section 4 describes the experimental setups and dataset preparation. Further, Section 5 presents performance analysis on various datasets under different dataset sizes and comparisons with state-of-the-art techniques. Section 6 covers the analysis highlighting various capabilities of PNN and ablation studies on PNN. The paper concludes in section 7 with a discussion on limitations and future scopes.

\section{Problem Formulation}\label{sec:prob}

We define the operation space $\mathcal{O}$ of given a rotating machine is defined in terms of different operation categories $O_i \in \mathcal{O}$ consisting of faulty operations $f_i \in \mathcal{F}$ and healthy operations $h_i \in {H}$. Thus, operation space $O$ can be represented as a union of Fault space $F$ and healthy operation space $H$, i.e., $\mathcal{O} = \mathcal{F} \cup {H}$. We assume that there are $N$ faulty operations and one healthy operation category. If $M$ numbers of observation vectors $x_k \in \chi $ of size $L$ where $M<<L$ are given, then our objective is to find a classifier $\mathcal{G}(.)$ such that overall classification cost $\mathcal{C}(\chi)$ is minimized as in \eqref{eqn:cost}. 

\vspace{-6.0ex}
\begin{equation}
    \Tilde{\mathcal{G}} =  \arg \min_{\mathcal{G}()} \mathcal{C}(\chi) =   \arg \min_{\mathcal{G}()} \sum_{k=1}^{M} \mathbf{1}(\mathcal{G}(x_k), O_k)
    \label{eqn:cost}
\end{equation}

Where, indicator function $\mathbf{1}(a,b) := 0$ if $a=b$, and $1$ if $a\neq b$. The $\mathcal{G}(x_k)$ is the predicted category by classification function $\mathcal{G}(.)$ for input $x_k$, and  $O_k \in O$ is true category for the observation $x_k$. Since the number of observations $M$ is very small compared to the subspace of observation $x \in  \mathbb{R}^L$, i.e., small-size database, the optimal $\Tilde{\mathcal{G}}$ depends upon the available dataset $\chi$. The optimal dataset-dependent classification function is denoted as $\Tilde{\mathcal{G_{\chi}}}$. Thus the problem can be formulated to estimate the dataset-dependent classifier  $\Tilde{\mathcal{G_{\chi}}}$ such that it is close to the global optimal classifier $\Tilde{\mathcal{G}}$ i.e. $\Tilde{\mathcal{G_{\chi}}} \simeq \Tilde{\mathcal{G}}$   

This work uses vibration data as observation data $X$ to find the optimal classifier $\Tilde{\mathcal{G}}$. A deep learning-based classifier (PNN) $\mathcal{H}$ has been proposed in this work. This classifier $\mathcal{H}$ maximizes the classification accuracy $Acc$ for the given small training set $M$, with inherent noise $N_s$ in the dataset by minimizing the classification error cost in \eqref{eqn:cost}. We discuss the proposed methodology in detail in the next section.

\begin{figure*}[!ht]
\label{fig:Arch}
\centering
\begin{center}
\includegraphics[width = 6in]{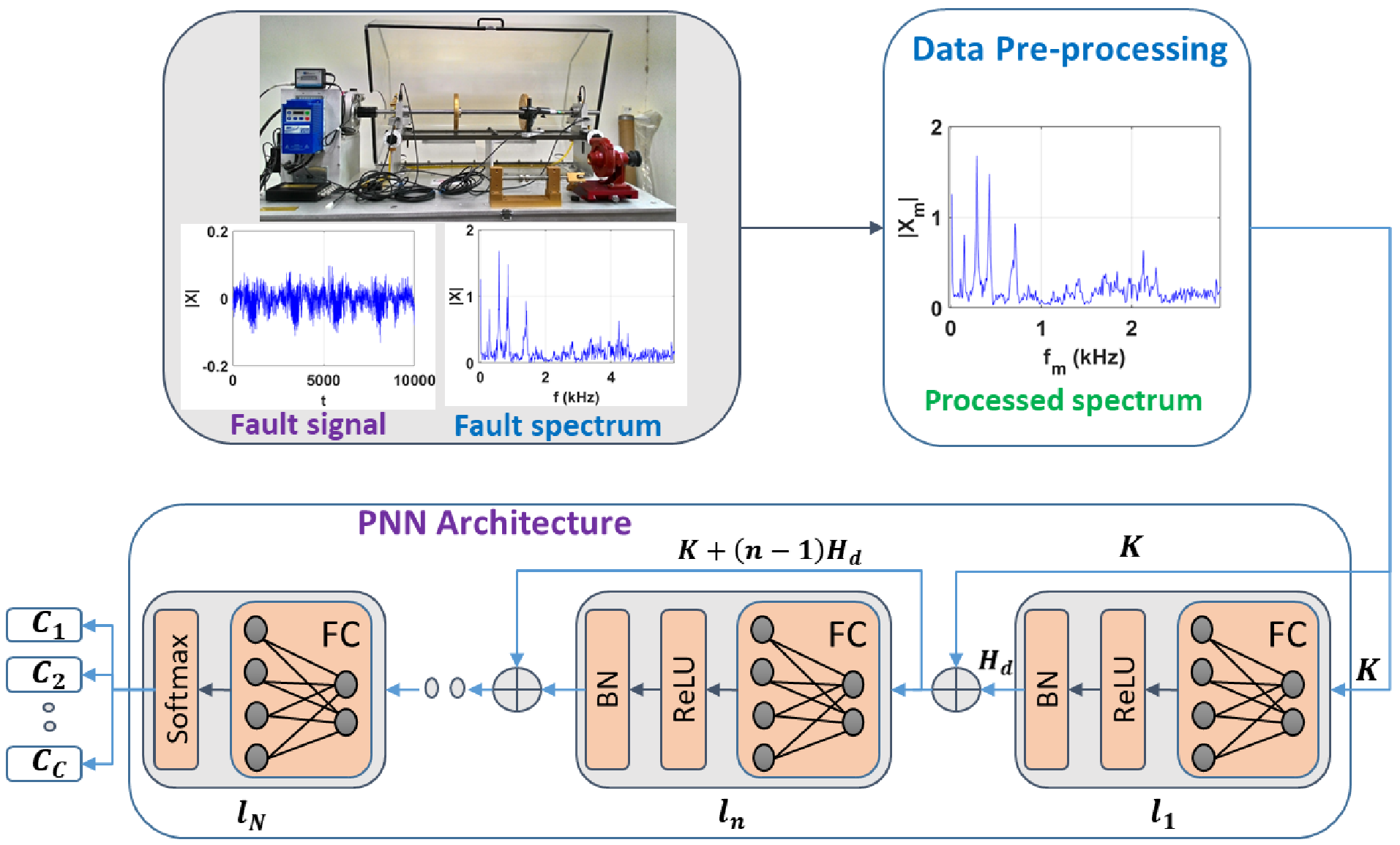}
\end{center}

\caption{An overview of the proposed unified fault detection technique: Original datasets of various sizes are standardized to a single size (16384), and then PNN of $N$ layers and $C$ number of fault classes is trained on this data. The first layer $(l_1)$ of PNN has input $X$ of size $K$, and the output $z_h$ of size $H_d$, The intermediate layer $(l_n)$ receives the input from layer $l_{n-1}$ (size $H_d$) along with input from previous layers $l_{n-2}$ (size $K+(n-2)H_d$) and the final layer gives output $C_i$ ($i^{th}$ fault class). Here $n$ is the layer number \& $N$ is depth of the PNN.}
\label{fig:PNNOverview1}
\end{figure*}

\section{Proposed Methodology}\label{sec:ProposedMethodology}
This section describes the data preprocessing steps and the design of the proposed Progressive Neural Network (PNN). The schematic diagram of the preprocessing block and the PNN block is shown in Figure~\ref{fig:PNNOverview1}. Detailed descriptions of these blocks are provided in the following subsections.

\subsection{Data preprocessing and size standardization}\label{sec:Datapre-processing}
In the preprocessing step, the input signal $x$ is transformed to emphasize the fault-characterizing features. These features are often complex and exhibit nonlinear relationships with the operating frequency of the rotary machine and its harmonics  \cite{c1, c35}. To capture these relationships, the signal $x$ is transformed into its frequency domain counterpart $X$ using the Fast Fourier Transform (FFT).

\begin{figure}[h]
\centerline{
\includegraphics[width=2.90cm,height=3.25cm]{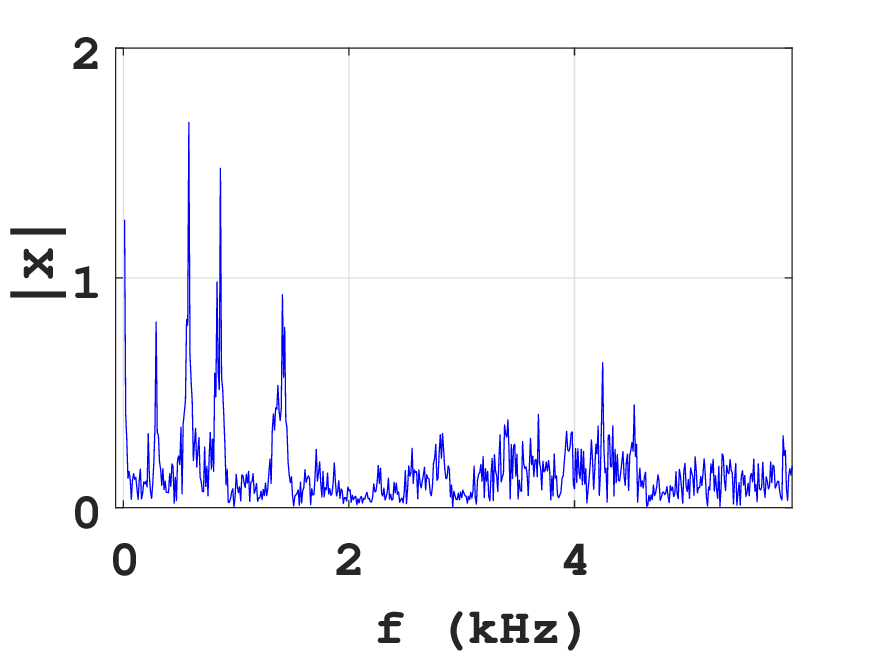}
\includegraphics[width=2.90cm,height=3.25cm]{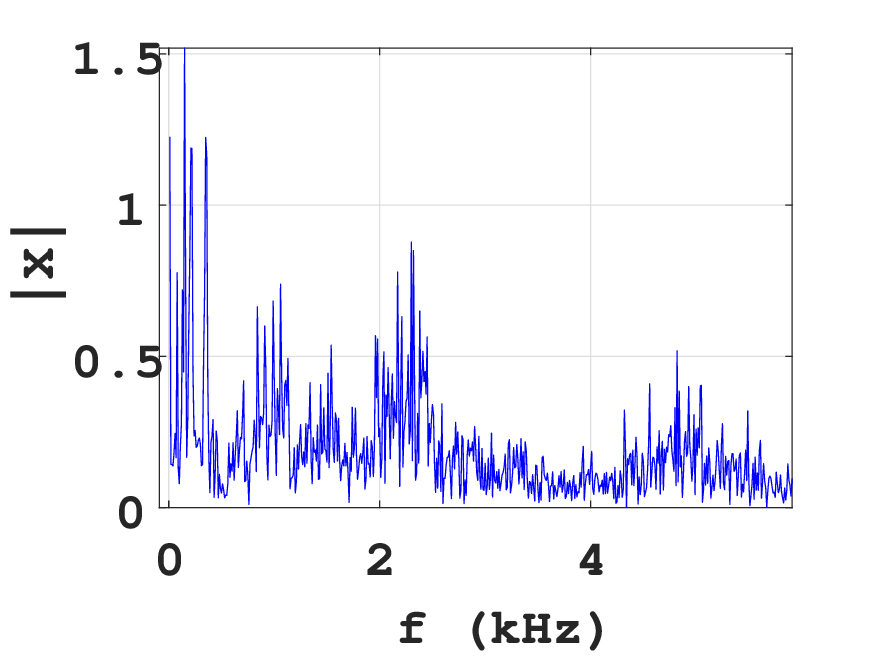}
\includegraphics[width=2.90cm,height=3.25cm]{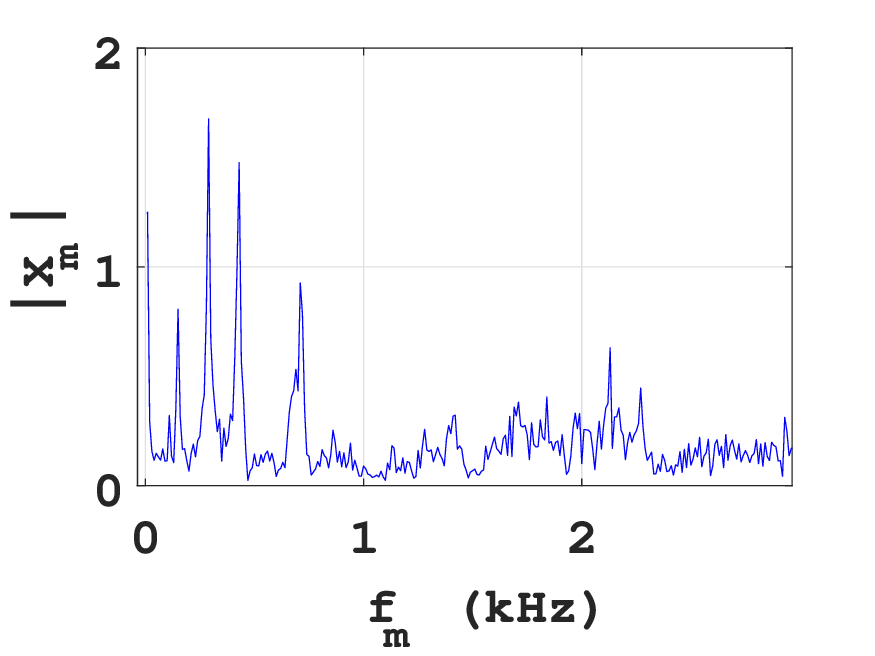}
}
\centerline{\footnotesize{ Original  \hspace{1.5cm}  N-point FFT  \hspace{1.5cm}  Max of bin }}
\caption{Typical IC-Engine data set fault spectrum with processing by N-point FFT (N=16384) and max of bin operation.}
\label{fig:Spectrums}
\end{figure}
Given that fault signals can vary in length and sampling rate, the transformed signal $X$ is standardized to ensure the alignment of features corresponding to the same harmonics. This is achieved by quantizing the signal to a fixed length of $N = 16384$ using the maximum value among the bins. This standardization process preserves the shape of the original spectrum $|X|$, unlike the direct application of a $16384$-point FFT on the input signal $x$, as illustrated in Fig.~\ref{fig:Spectrums}.

As shown in Fig.~\ref{fig:Spectrums}, the relative positions of peaks and their magnitudes, which represent the key features of the data, are well preserved after the proposed preprocessing. In contrast, the spectrum obtained from a direct $16384$-point FFT lacks this consistency. By retaining these critical features, the standardization step enhances the generalizability of the proposed classification framework, namely the Progressive Neural Network (PNN) architecture, across different types and sizes of fault datasets. The PNN architecture is discussed in detail in the following subsection.

\subsection{PNN architecture}\label{sec:PNNArchitecture}

The DNN, in general, has a large number of parameters, and due to this, to learn features from data, they need a large amount of training data. Our aim is to make a model for Industry 4.0, where the model size is small so that the small number of parameters can be turned with a small size of training data. 

\par We propose a novel Progressive Neural Network (PNN) architecture for fault classification, which progressively estimates and concatenates the higher-order features as shown in Fig. \ref{fig:PNNOverview1}. In the Figure, the input $X$ is of size $K$, the output $z_h$ of a PNN layer is of size $H_d$, and $C$ is the number of fault classes. The output of the final layer is the fault class $C_i$ corresponding to the input $x$. 
Our approach draws inspiration from the Fourier Neural Operator Deep Neural Network (DNN) introduced in \cite{c56}, where each layer processes the previous layer's output using a Fourier transform, a linear transform, and an inverse Fourier transform. The resulting output is concatenated with a locally linear-transformed version of the previous layer’s output. In contrast, our PNN architecture directly utilizes the Fourier spectrum as input, eliminating the need to process the output of preceding layers. This design significantly reduces the complexity of the DNN, decreases the number of processing nodes, and shortens the training time.

\begin{figure*}[]
\centering
\begin{center}
\includegraphics[width = 7in]{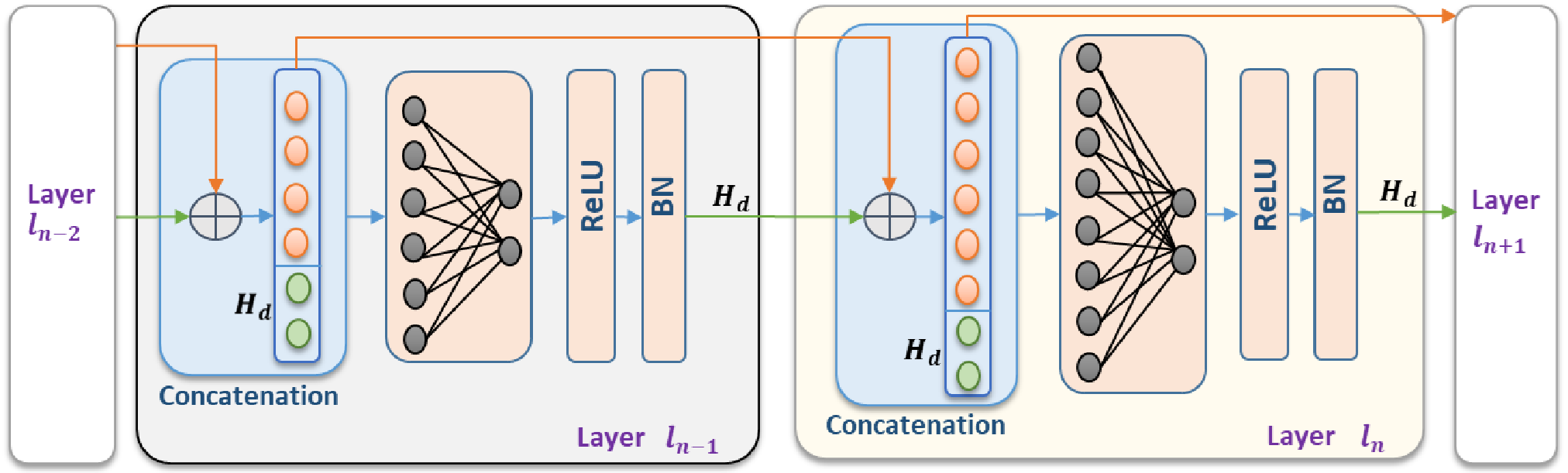}
\end{center}
\caption{Detailed design of a PNN processing blocks}
\label{fig:PNNBlock}
\end{figure*}

\par Figure \ref{fig:PNNBlock} illustrates the detailed structure of a PNN layer. Each PNN layer, denoted as $l_n$, is a fully connected layer with a progressively increasing input size while maintaining a fixed output size of $H_d$. In this architecture, the output of each layer is formed by stacking the input and output of the previous layer. Specifically, the first layer, $l_1$, takes the original input $X$ of size $K$. Subsequent layers, such as $l_n$, progressively combine the original input $X$ (of size $K$) with the outputs of all preceding layers (of total size $(n-1)H_d$), as depicted in Figure \ref{fig:PNNBlock}.

So, the next layer receives both the original data and the features learned from previous layers. Due to this, the feature extraction at each layer becomes better and better. At the time of back-propagation, a direct path is available for gradient flow from the last layer to the first layer \& due to this flow, the problem of the vanishing gradients does not appear.  
\par The activation $a^n$ at layer $l_n$ is computed as given in \eqref{eqn:actvn} for output $z_h^n$ of a layer $l_n$. The update in output $z_h^n$ for weight $w$ and bias $b$ with input $X$ and $a^n$ is done as in  \eqref{eqn:output}. 

\newcommand\doubleplus{+\kern-1.3ex+\kern0.8ex}
\begin{equation}
\left.
 \begin{array}{ c l }
  a^{n}&=\sigma(z^{n}_h) \doubleplus a^{n-1}  \quad \textrm{if  } n > 1 \\
   a^{n}&=\sigma(z^{n}_h) \qquad \qquad \textrm{o.w.}
 \end{array}
 \right\}
\label{eqn:actvn}
\end{equation} 

Where $\sigma(.)$ is the activation function, and the symbol \doubleplus represents the concatenation.

\begin{equation}
\left.
 \begin{array}{ c l }
  z^{n}_h&=w^{n}a^n+b^{n}  \quad \textrm{if  } n > 1 \\
  z^{n}_h&=w^{n}X+b^{n} \qquad \textrm{o.w.}
 \end{array}
\right\}
\label{eqn:output}
\end{equation}

\par Also, the PNN does not suffer from the problem of vanishing gradient due to direct connections of each layer with the input layer. Thus, the PNN can have a large depth. The PNN has been tested with more than six layers, and the obtained performance is far better than the vanilla DNN network. The architecture of the PNN-based model for a depth of $N$ layers is as given in \eqref{eqn:layer}.

\begin{equation}
\resizebox{0.9\columnwidth}{!}{$
\left.
 \begin{array}{ c l }
&\text{Layer}[l_1]: \text{Linear}(K, H_d), \text{ReLU,BN}  \\
&\text{Layer}[l_n]:  \text{Linear} (K+(n-1)H_d, H_d), \text{ReLU,BN}  \\
&\text{Layer}[l_N]: \text{Softmax}(\text{Linear} (K+(N-1)H_d, C)) 
 \end{array}
\right\}
$}
\label{eqn:layer}
\end{equation}

Here, BN represents the 1D batch normalization. As given in the PNN architecture, the hidden layer size $H_d$ is constant for all layers, and $(H_d<<K)$. So, the increment in the number of neurons in successive layers is by a constant amount of $H_d$. Due to this, the number of parameters in the PNN is very small compared to Vanilla DNN (VDNN) based models of the same depth. The number of parameters in a typical VDNN of six layers will be $~0.67K^2$, with each successive layer having only $50\%$ neurons of the previous layer. But for the six layers of PNN, the number of parameters will be $6KH_d+15H_d^2$, where $H_d$ is the hidden layer's size, which is very small. In the next section, we describe the experimental setup and datasets on which the performance of PNN has been evaluated.

\section{Experimental Setup, Dataset Collection, and Simulation}\label{sec:ExperimentalSetups}
\setlength{\tabcolsep}{6pt}
\renewcommand{\arraystretch}{1.1}

\begin{table}[h!]
\centering
\caption{Datasets and their descriptions; $C$: Number of fault classes, $M$: total samples. Details of open-source datasets are available in the referenced literature \cite{c2,c3}.} 
\label{tab:alldatasets}

\resizebox{\columnwidth}{!}{
\begin{tabular}{|c|l|l|c|c|}
\hline
S.No & Dataset & Description & $C$ & $M$ \\
\hline
1 & CWRU & Open-source bearing dataset from CWR University & 10 & 100 \\
\hline
2 & PB & Open-source bearing dataset from Paderborn University & 3 & 560 \\
\hline
3 & UoC & Open-source gear fault dataset from University of Connecticut & 9 & 936 \\
\hline
4 & JNU & Open-source bearing dataset from Jiangnan University & 12 & 240 \\
\hline
5 & MFPT & Fault dataset from Society for Machinery Failure Prevention Technology & 15 & 186 \\
\hline
6 & SEU & Gearbox fault dataset from SEU Gearbox Dataset & 20 & 100 \\
\hline
7 & IC-Engine (ICE) & In-house IC-engine dataset from an industrial manufacturer \cite{c10} & 7 & 405 \\
\hline
8 & Simulator (Sim) & In-house dataset from SpectraQuest Machinery Fault Simulator & 7 & 525 \\
\hline
\end{tabular}
}
\end{table}

This section describes the experimental setup, dataset collection, and simulation procedures. To simulate small dataset scenarios, we divided the datasets into training and testing sets with ratios of $10\%-90\%$ (very small dataset case) and $25\%-75\%$ (small dataset case). Table~\ref{tab:alldatasets} summarizes the details of the datasets, where $C$ denotes the number of fault classes, and $M$ is the total number of samples across all classes. 

\par
\begin{figure}[h!]
  \centering
   \includegraphics[width=\columnwidth]{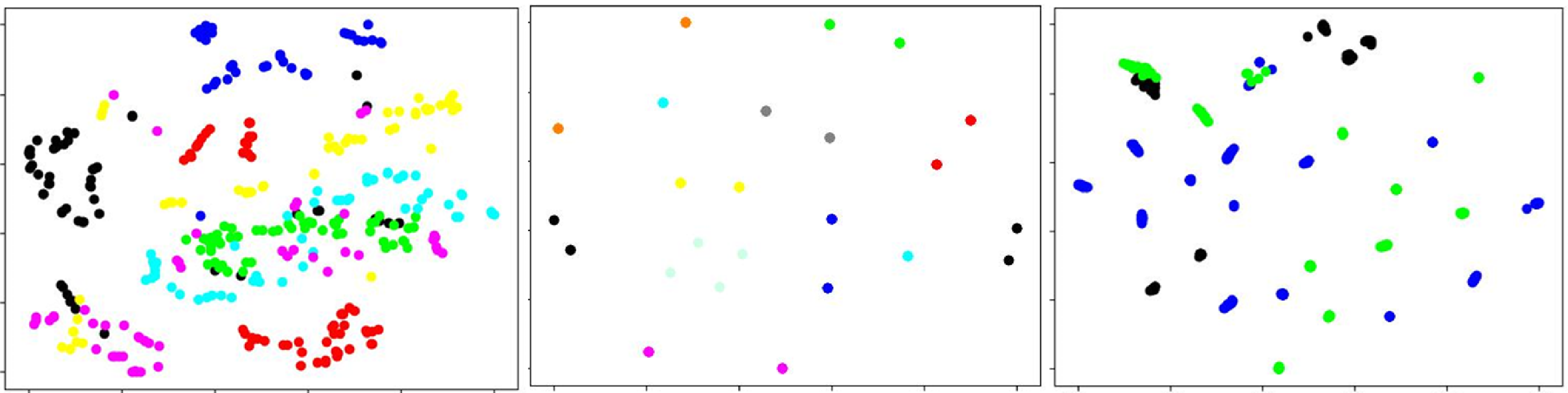}
   \centerline{\footnotesize{ ICE  \hspace{1.5cm}  CWRU  \hspace{1.99cm}  PB }}
  \caption{t-SNE plot of fault classes for \emph{ICE}, \emph{CWRU}, and \emph{PB} datasets. Each color represents a fault type in the plot.}
  \label{fig:DistributionOfEngineCWRUPB}
\end{figure}

 From Table~\ref{tab:alldatasets}, it is evident that the datasets contain very limited samples per fault class, and the distribution of samples across fault classes is highly imbalanced. For instance, in the JNU dataset, there are only $2$ samples for $12$ fault classes. Datasets 1 through 6 are widely recognized as standard benchmarks in fault detection and classification research, with their descriptions and prior results available in \cite{c2,c3}. Figure~\ref{fig:DistributionOfEngineCWRUPB} depicts the t-SNE projection of fault features for the \emph{IC-Engine} (ICE), \emph{CWRU}, and \emph{PB} datasets. It is evident from the figure that the fault features are significantly mixed and lack clear separability, making fault classification a highly challenging task.
\par In the following subsections, we describe the in-house experimental setups used for fault simulation, including the \emph{SpectraQuest} machinery fault simulator and the real fault data captured from the \emph{ICE test-rig}.

\subsection{Machinery fault simulator test-bed}\label{sec:MachineryFaultSimulator}

\begin{figure}[!ht]
\centering
\begin{center}
\includegraphics[width=3.35in, height=2.150in]{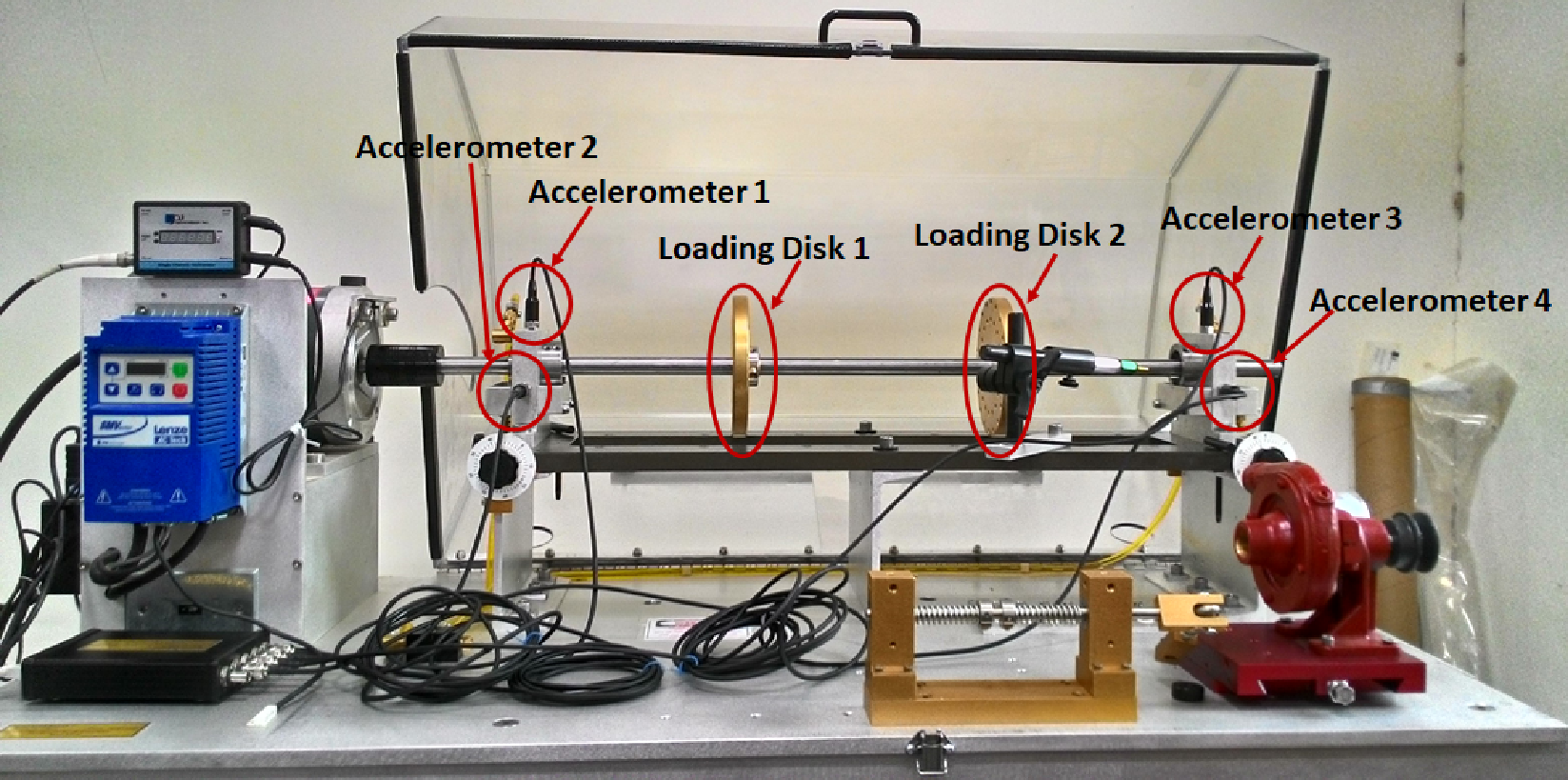}
\end{center}
\caption{\emph{SpectraQuest} machinery fault simulator with components.}
\label{fig:faultsim}
\end{figure}

The machinery fault simulator test-bed from \emph{SpectraQuest}, shown in Fig.~\ref{fig:faultsim}, is designed to seed various types of faults on a rotating shaft under load. It is equipped with four single-axis accelerometer sensors (\emph{ICP SQI608A11}), mounted at different positions to capture vibration data, as illustrated in Fig.~\ref{fig:faultsim}. The rotational speed of the shaft is controlled using a variable frequency speed controller attached to a motor, which drives the shaft.

\par For the experiments, the shaft's rotational speed was maintained at $20\,\text{Hz}$, and the vibration signals were sampled at a rate of $51\,\text{kHz}$. Vibration data were recorded for six different simulated fault types and one normal operating condition. For each fault class, $75$ samples were collected from each sensor position. Detailed descriptions of the simulated faults are provided in Table~\ref{tab:simdataset}.

\setlength{\tabcolsep}{6pt} 
\renewcommand{\arraystretch}{1} 
\begin{table}[]
\centering
\caption{Types of fault seeded and their description for machinery fault simulator test-bed } 
\label{tab:simdataset}
\renewcommand{\arraystretch}{1.5} 
  \resizebox{\columnwidth}{!}{%
\begin{tabular}{ |c | l | l |}
\hline
{S.No} & Fault Type &  {Fault Description}\\
\hline
0 & Dynamic Couple & \begin{tabular}[c]{@{}l@{}}Both loading disk have mass in opposite side \end{tabular} \\
\hline
1 & Unbalanced Mass & \begin{tabular}[c]{@{}l@{}}Only one loading disk has mass \end{tabular} \\
\hline
2 & Lateral Misalignment & \begin{tabular}[c]{@{}l@{}}Both end of shaft are at different displacement \end{tabular} \\
\hline
3 & Bent in Shaft & \begin{tabular}[c]{@{}l@{}}Shaft with bent is used with loading disks \end{tabular} \\
\hline
4 & Crack in Shaft & \begin{tabular}[c]{@{}l@{}}Shaft with crack in the middle is used \end{tabular} \\
\hline
5 & Faulty Bearing & \begin{tabular}[c]{@{}l@{}}Faulty bearings were used at both ends \end{tabular} \\
\hline
6 & Healthy Motion & \begin{tabular}[c]{@{}l@{}}Motion without any seeded fault \end{tabular} \\
\hline
\end{tabular}
}
\end{table}

\subsection{IC-Engine Test-Rig}\label{sec:IC-EngineTest-Rig}

We have utilized data from a single-cylinder internal combustion engine (IC-Engine) test rig \cite{c10} of a commercial two-wheeler manufacturing company to record the vibration signals from the engine. This setup has a single-cylinder internal combustion engine with an optical encoder for speed measurement. Four sensors were placed in four different positions on the engine to record vibration data from the engines. The engine's rotation speed for testing was kept at $2500$ RPM, and the recorded signals from sensors were sampled at $50kHz$. The vibration data is recorded for six different types of seeded faults and one healthy engine operation. The types of fault seeded and their description are given in Table \ref{tab:enginedataset}.  This dataset has an uneven distribution of the samples with a total of $40$ samples for the training with $7$ fault classes.

\setlength{\tabcolsep}{6pt} 
\renewcommand{\arraystretch}{1} 
\begin{table}[]
\centering
\caption{Types of fault seeded in IC-Engine test-rig} 
 \label{tab:enginedataset}
\renewcommand{\arraystretch}{1.1} 
  \resizebox{\columnwidth}{!}{ %
\begin{tabular}{|c|l|c|l|}
\hline
{S.No} & Fault Type &  Samples & {Fault Description}\\
\hline
0 & {PGW}  & {64} & \begin{tabular}[c]{@{}l@{}}Primary gear whining, misalignment of gears \end{tabular} \\
\hline
1 & {MRN} & {65} & \begin{tabular}[c]{@{}l@{}}Magneto rotor noise, pulsar coil starts rubbing with other \\ parts due to reduced gap. \end{tabular} \\
\hline
2 &  \begin{tabular}[c]{@{}l@{}}{TAPPET}\\(TPT)\end{tabular} & {59} & \begin{tabular}[c]{@{}l@{}}Tappet noise, high deviation of tappet clearance from \\ ideal settings \end{tabular} \\
\hline
3 & {CHN} & {40} & \begin{tabular}[c]{@{}l@{}}Cylinder head noise, inappropriate setting of top dead\\ center causing cylinder head slapping.  \end{tabular} \\
\hline
4 & {PGD} & {57} & \begin{tabular}[c]{@{}l@{}}Primary gear damage noise, abnormality in drive gear,\\ driven gears assembly in the form of tooth damage,\\ tooth profile error, and inclined bore. \end{tabular} \\
\hline
5 & {CCN} & {60} & \begin{tabular}[c]{@{}l@{}}Cam chain noise, noise from  Cam chain stress.\end{tabular} \\ 
\hline
6 & \begin{tabular}[c]{@{}l@{}}Healthy\\ Motion \\(HEM)\end{tabular} & {60} & \begin{tabular}[c]{@{}l@{}}Motion without any seeded fault. \end{tabular} \\
\hline

\end{tabular}
}
\end{table}

\section {PNN Performance Analysis}\label{sec:PNNPerformanceAnalysis}

In this section, we evaluate the proposed Progressive Neural Network (PNN)-based fault classification framework for its classification performance across various datasets. The results are compared with both classical and state-of-the-art techniques to demonstrate the effectiveness of the proposed approach. 

\par As part of the performance analysis, in our experiments, we trained the PNN-based models of hidden layer size $100$ and depth $6$ with a learning rate of 0.0001 using the Adam optimizer. Training was performed over 30 epochs with a batch size of 8. Weight decay was set to $0.0001$, with Cross-entropy loss. Hyperparameters were selected for \emph{ICE} dataset, based on grid search within the following ranges: learning rate/Weight decay [0.0001, 0.01], batch size [4, 16],  hidden size [10,1000], and PNN depth [3,6].

\subsection{Performance Comparison with Classical ML and DL Techniques}\label{sec:ClassicalMLDL}

The proposed PNN-based framework is fundamentally a deep neural network (DNN). Hence, it is natural to compare its performance with that of widely used DNN architectures. We evaluated the PNN against classical machine learning (ML) such as Support Vector Machines (SVM)  \cite{c40}, and deep learning (DL) techniques, including Vanilla DNN (VDNN) \cite{c12}, and advanced DL models such as ResNet18 \cite{c38}, VGG11 \cite{c39}, LeNet \cite{c42}, AlexNet \cite{c45}, and LSTM \cite{c43}.

\par The PNN and VDNN models with six layers (PNN6 and VDNN6), along with SVM using the RBF kernel \cite{c40}, were tested on all datasets with training-to-testing ratios of $10$-$90\%$, $25$-$75\%$, and $75$-$25\%$. The best accuracy for each dataset and model configuration was recorded for comparison.

\par Additionally, since the PNN can function as a fully connected (FC) layer in deep learning models, we evaluated the performance of ResNet18 and VGG11 with PNN6 as the FC layer. These configurations, referred to as ResNet18+PNN6 (RPNN) and VGG11+PNN6 (VPNN), were tested under similar conditions.

\par Table \ref{tab:ModelSizesAccAll} summarizes the performance of these models, including PNN6 as the FC layer in DL models. It is evident from the table that PNN-based models achieve significantly better accuracy across all division ratios and datasets compared to VDNN and SVM-based models. The performance of RPNN and VPNN configurations also surpasses that of VDNN and SVM. However, VPNN exhibits slightly inferior performance compared to PNN6 and RPNN due to the vanishing gradient problem associated with VGG-based models, which hinders their classification accuracy in scenarios with limited training samples.

\par Notably, VDNN and SVM models fail to outperform a random classifier across all datasets and division ratios. Despite challenges in single-shot or few-shot learning scenarios, such as the $10$-$90\%$ division ratio for the SEU, MFPT, and JNU datasets (where the number of samples per class is minimal $1$ or $2$), PNN models still demonstrate superior performance compared to VDNN and SVM. 

The performance of the PNN6 models as classifiers across all datasets, under various training-testing division ratios (DR), is evaluated using the Area Under the Receiver Operating Characteristic Curve (AUROC), F1 Score, and standard deviation (SD). Table \ref{tab:AUROC} summarizes these metrics. For the $75$-$25\%$ division ratio, where the classifiers are provided with sufficient training data, the models achieve an AUROC of nearly $100\%$ with a comparable F1 Score across all datasets. In this scenario, the SD values are minimal, reflecting consistent model performance.

The PNN6-based models consistently exhibit high AUROC and F1 Score values across all division ratios and datasets, highlighting their superior classification accuracy. Although the SD indicates variability, it remains relatively small, suggesting stable performance during repeated runs.

For smaller division ratios, such as $10$-$90\%$, the performance decreases due to limited training data. However, the metrics remain competitive, and the variation remains within acceptable limits. Notably, datasets like JNU, MFPT, and SEU are specific cases where the F1 Score is very poor due to the training samples being predominantly of one or two samples per class.
\par
\begin{figure}[!ht]
\centering
\begin{center}
\includegraphics[width=3.35in, height=2.150in]{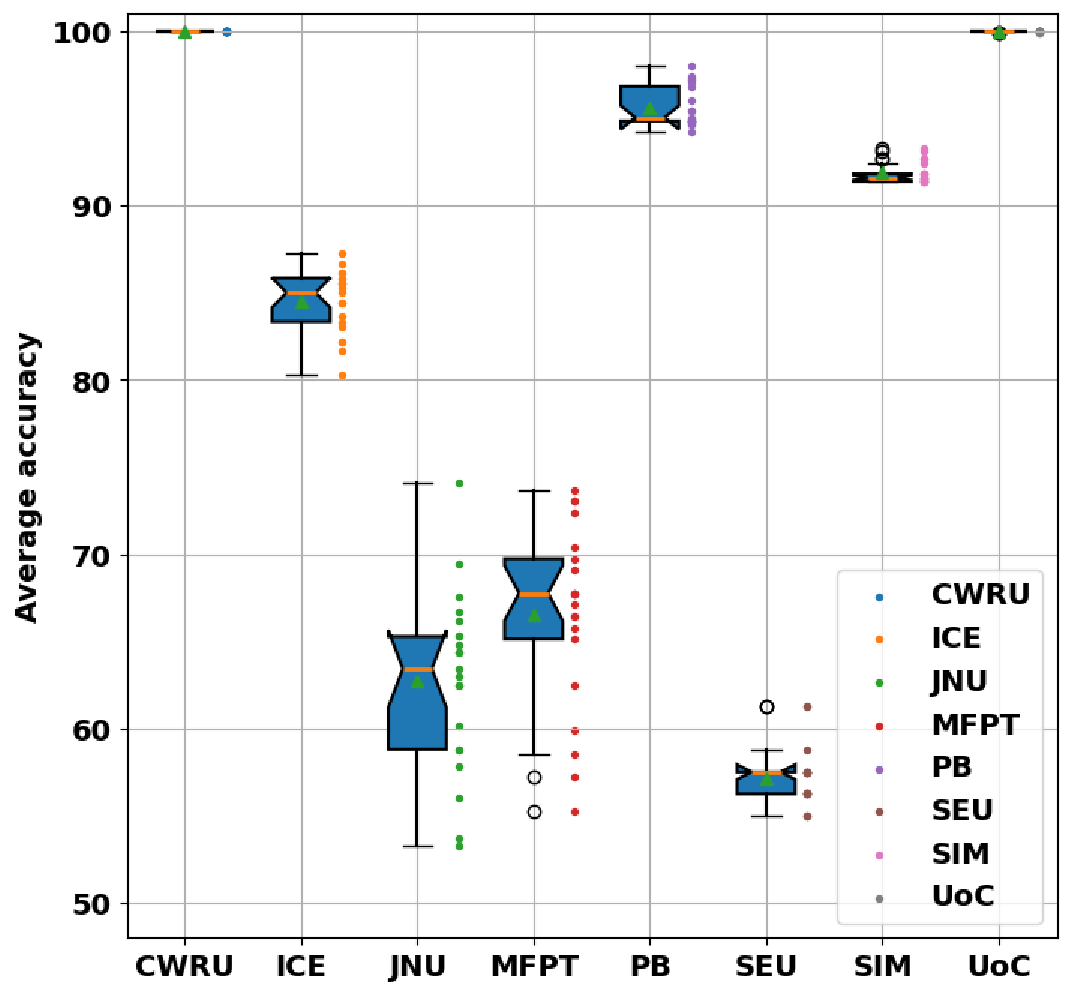}
\end{center}
\caption{\emph{PNN6} classification performance on all datasets for 10-90\% division ratio.}
\label{fig:Box90}
\end{figure}

 Figure~\ref{fig:Box90} illustrates the classification performance of PNN6 across all datasets for a small training dataset division ratio of $10$-$90\%$. Under this scenario, the performance is suboptimal for certain datasets, exhibiting significant deviations and low classification accuracy. For complex datasets such as JNU and MFPT, the limited training data further results in poor accuracy and large deviations. Conversely, when a larger training dataset is employed ($75$-$25\%$ division ratio), the classification performance exceeds 90\% with minimal deviation, as highlighted in Table~\ref{tab:AUROC}.

\setlength{\tabcolsep}{6pt} 
\renewcommand{\arraystretch}{1.0} 
\begin{table}[h!]
\centering
\tiny
\caption{Comparison of accuracy (in \%) for PNN6 (PNN of depth 6) with VDNN6, SVM (RBF kernel), Resnet18+PNN6 (RPNN), VGG11+PNN6 (VPNN) for $10-90\%$ (10\%), $25-75\% 
 (25\%)$ and $75-25\% (75\%)$ training-testing division ratios (DR).}
 \label{tab:ModelSizesAccAll}
\renewcommand{\arraystretch}{1.95} 
  \resizebox{\columnwidth}{!}{%
\begin{tabular}{|c |c | c | c |c|c | c | c |c|c|}
\hline
\multicolumn{2}{|c|}{Model type}  & \multicolumn{8}{|c|}{Accuracy (in \%)} \\
\cline{3-10}
\multicolumn{2}{|c|}{DR} & CWRU  & ICE & JNU & MFPT & PB & SEU & Sim & UoC \\
\hline
\multirow{3}{*}{ \rotatebox[origin=c]{90}{\parbox[c]{1cm}{\centering SVM }}} &10\% & 46.3 & 14.76 & 27.67 & 6.4 & 43.0 & 43.6 &
48.4 & 15.26 \\
\cline{2-10}
&25\% & 46.67 & 15.7 & 20.1 & 9.4 & 43 &
45.8 & 23.7 & 21.6 \\
\cline{2-10}
&75\% & 44.6 & 15.3 & 19.0 & 10.6 & 43.3 &
62.5 & 16.1 & 22.3 \\
\hline
\multirow{3}{*}{ \rotatebox[origin=c]{90}{\parbox[c]{1cm}{\centering VDNN6 }}} &10\% & 52.0 & 24.3 & 23.6 & 40.1 & 41.7 &
32.0 & 23.7 & 20.9 \\
\cline{2-10}
&25\% & 53.2 & 25.4 & 24.7 & 42.6 & 42.8 & 31.3
& 25.8 & 22.9 \\
\cline{2-10}
&75\% & 72.5 & 33.3 & 35.7 & 26.7 & 47.5 & 54 &
31.7 & 39.2 \\
\hline
\multirow{3}{*}{ \rotatebox[origin=c]{90}{\parbox[c]{1cm}{\centering RPNN }}} & 10\% & \textbf{95} & {87.2} & \textbf{57} & {57.5} & {92} & \textbf{82.2} & 90.8 & {99.9} \\
\cline{2-10}
&25\% & \textbf{96.2} & {98} & {56.8} & 90.6 & \textbf{98.6} & {95.6} &  \textbf{100}&
\textbf{100} \\
\cline{2-10}
&75\% & 100 & {99.8} & {99.1} & 98.75 & {97.4} & \textbf{98.7} & \textbf{100} &
\textbf{100} \\
\hline
\multirow{3}{*}{ \rotatebox[origin=c]{90}{\parbox[c]{1cm}{\centering VPNN }}} & 10\% & 61 & \textbf{94.3} & {24.7} & 18.7 & {91.4} & {19.8} & 99.6 &
\textbf{100} \\
\cline{2-10}
& 25\% & 64.2 & \textbf{100} & {97.4} & 90.5 & {92.2} & {87.2} &  \textbf{100}&
\textbf{100} \\
\cline{2-10}
& 75\% & 96.2 & \textbf{100} & \textbf{100} & \textbf{100} & {98.6} & {95.6} & \textbf{100} &
\textbf{100} \\
\hline
\multirow{3}{*}{ \rotatebox[origin=c]{90}{\parbox[c]{1cm}{\centering PNN6}}} & 10\% & 100 & {93.6} & {55.0} & \textbf{61.0} & \textbf{97.3} & {60.6} & {99.2} & {99.9} \\
\cline{2-10}
&25\% & 100 & 96.95  & \textbf{98.6} & \textbf{100} & \textbf{98.6} & 97.7 &
{99.74} & \textbf{100} \\
\cline{2-10}
&75\% & \textbf{100} & \textbf{100} & \textbf{100} & \textbf{100} & \textbf{98.6} & {95.2} & \textbf{100} & \textbf{100} \\
\hline
\end{tabular}
}
\end{table}

\setlength{\tabcolsep}{4pt} 
\renewcommand{\arraystretch}{1} 
\begin{table}[h!]
\centering
\tiny
\caption{PNN6 model performance across datasets, evaluated with Accuracy (mean ± SD), F1-score (mean), and AUROC (mean) for 50 iterations and varying training-testing division ratios (10-90\%, 25-75\%, 75-25\%). The abbreviations are as follows: [Accuracy: Acc], [Division Ratio: DR], [Dataset: DS], [CWRU: CW], and [MFPT: MF].}
\label{tab:AUROC}
\begin{tabular}{|c|c|c|c|c|c|c|c|c|c|}
\hline
 DR           & \multicolumn{3}{|c|}{10-90\%}                   & \multicolumn{3}{|c|}{25-75\%}                   & \multicolumn{3}{|c|}{75-25\%} \\
\cline{2-10}
DS & Acc (\%) & F1   & AU & Acc (\%) & F1   & AU & Acc (\%) & F1   & AU                \\
\hline
CW & 100$\pm$0 & 100   & 100   & 100$\pm$0 &   100  & 100   & 100$\pm$0  & 100   & 100    \\
ICE  & 84.5$\pm$1.8 & 84.5    & 91.3  & 93.5$\pm$1.0   & 93.5    & 95.7   & 99.1$\pm$0.4   & 99.1    & 99.5     \\
JNU & 62.8$\pm$5.0 & 62.8    & 80.2   & 92.4$\pm$1.6   & 92.4    & 96.1   & 99.9$\pm$0.4   & 99.9    & 100     \\
MF &  66.5$\pm$5.1 &66.5    & 81.7   & 86$\pm$3.0   & 86.0    & 92.9    & 100$\pm$0  & 100  & 100     \\
PB &  95.6$\pm$1.1 & 95.6    & 96.6   & 98.4$\pm$0.1   & 98.4    & 98.8   & 98.8$\pm$0.4   & 98.8    & 99.0    \\
SEU  & 57.1$\pm$1.6 & 57.1    & 81.2  & 56.9$\pm$1.8    & 56.9    & 81.0   & 97.5$\pm$0.8   & 97.5    & 98.9     \\
Sim  & 91.9$\pm$0.6& 91.9    & 95.4   & 99.3$\pm$0.1   & 99.3    & 99.5   & 100$\pm$0     & 100   & 100   \\
UoC &  100$\pm$0 & 100   & 100   & 100$\pm$0  & 100   & 100  & 100$\pm$0  & 100   & 100   \\
\hline
\end{tabular}
\end{table}

\setlength{\tabcolsep}{6pt} 
\renewcommand{\arraystretch}{1} 
\begin{table}[h!]
\centering
\caption{Comparison of proposed PNN-based technique for $80-20\%$ division ratio with accuracy reported in DL-based benchmark study paper \cite{c3}.  The accuracy of multiple techniques is average accuracy in \%. }
 \label{tab:Benchmark}
\renewcommand{\arraystretch}{1.1} 
  \resizebox{\columnwidth}{!}{%
\begin{tabular}{ |c | c | c | c |c|c| c | c |}
\hline
\multirow{2}{*}{Dataset }&\multicolumn{7}{c |}{Accuracy (in \%) } \\
 \cline{2-8} 
 & AEC&   CNN&  LeNet&  AlexNet&  ResNet18&  LSTM  &  PNN6  \\
\hline
CWRU  & \textbf{100} & 99.85&  \textbf{100}&  \textbf{100}&  99.77&  99.92  & 100\\
\hline
JNU  & 95.77 &  92.88&  95.05 & 95.37 & 96.49&  95.23 & \textbf{100}\\
\hline
MFPT & 94.95 &   79.96 & 93.75&  92.04&  92.27 & 93.09& \textbf{100}\\
\hline
PB & 74.62  &     90.57 & 95.85 & 95.18& \textbf{98.77} & 93.86 & 98.6\\
\hline
SEU &  96.71  &   96.86 & 98.09 & 97.2 & \textbf{99.85} & 97.35 & 95.2\\
\hline
UoC & 92.53  &   65.36 & 83.38& 75.04 & 88.13&  80.82 & \textbf{100}\\
\hline
\begin{tabular}[c]{@{}l@{}}Mean\\Accuracy\end{tabular} & 92.43  &   87.58 & 94.35& 92.47 & 95.88&  93.38 & \textbf{98.9}\\
\hline
\end{tabular}
}
\end{table}

\begin{figure}[!ht]

\begin{tikzpicture}[tight background]
\begin{axis}[
width=3.5in,
    height=2.250in,
    ylabel = {[Accuracy, AUROC]},
     xlabel=Division Ratio (\%),
     xmin=0,
     xmax=8,
     xtick={0,1,2,3,4,5,6,7,8},
     xticklabels={[10-90],[20-80],[30-70],[40-60],[50-50],[60-40],[70-30],[80-20],[90-10]},
     x tick label style={font=\tiny},
     y tick label style={font=\tiny},
     ymajorgrids=true,xmajorgrids=true,
grid style=dotted,legend columns=4,
legend style={font=\footnotesize,  at={(0.62, 0.10)}, anchor=south},
scale=1.0
]

\pgfplotstableread{ICEAccAllDiv.dat}{\pistonkinetics}
\addlegendentry{Accuracy}  
\addplot [green,mark=*, line width=1.0pt] table [x={Iter}, y={Acc}] {\pistonkinetics};

\addlegendentry{AUROC}  
\pgfplotstableread{ICEAUROCAllDiv.dat}{\pistonkinetics}
\addplot [blue,mark=o, line width=1.0pt] table [x={Iter}, y={AUROC}] {\pistonkinetics};
    
\end{axis}
\end{tikzpicture}
\caption{PNN6 performance in terms of Accuracy (\%) and AUROC, with different training-testing division ratios for ICE dataset. The division ratio of [10-90\%] is a few-shot or very small-size dataset scenario, and the division ratios [20-80\%] \& [30-70\%] are small-size dataset scenarios.}
\label{fig:ICEAccF1AUROC} 
\end{figure}
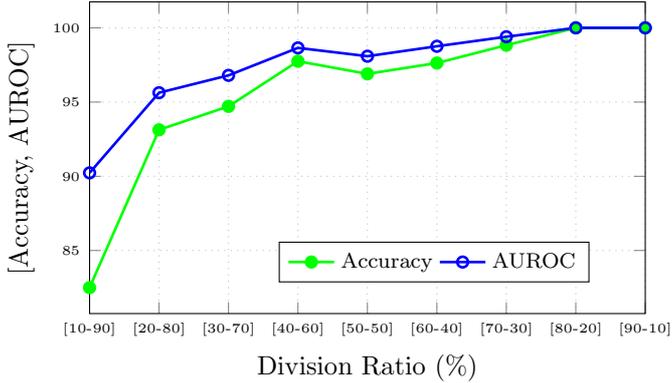

The CWRU dataset is relatively simple (as illustrated by the t-SNE plot in Figure~\ref{fig:DistributionOfEngineCWRUPB}). As a result, the performance of PNN6 remains high even with a small training dataset. However, as the model complexity increases (e.g., in the cases of RPNN and VPNN), the performance decreases due to the higher requirement for larger training datasets. Figure~\ref{fig:ICEAccF1AUROC} depicts the performance of PNN6 on the ICE dataset in terms of accuracy and AUROC across multiple dataset division ratios. The results show a consistent improvement in performance as the size of the training dataset increases.
\par Furthermore, we compared the proposed PNN framework with advanced DL models as evaluated in benchmark \cite{c3}. Table \ref{tab:Benchmark} demonstrates the accuracy achieved by various DL models on six open-source datasets. Despite being a relatively smaller model, PNN achieves accuracy comparable to or better than the benchmark models, with a mean accuracy that outperforms all evaluated deep models.

\subsection{Performance Comparison with State-of-the-Art Techniques}\label{sec:LatestTechniques}

\par Table~\ref{tab:OtherTech} provides a comprehensive performance comparison between the proposed PNN6 model and several state-of-the-art deep learning techniques. The listed models predominantly leverage deep neural network architectures, with many incorporating transfer learning (TL) to enhance their classification capabilities.

\par For smaller dataset sizes, the FDFSL technique \cite{c7}, which employs a Siamese CNN-based network for few-shot learning (FSL), achieves competitive performance on the CWRU dataset with six training samples per fault class. However, its accuracy falls short when compared to PNN6 at the $10$-$90\%$ division ratio. Similarly, TabPFN \cite{c48}, a transformer-based model trained on a vast corpus of synthetic data and fine-tuned on target datasets, performs slightly worse than PNN6 at $10\%$ and $25\%$ division ratios. Notably, TabPFN's dependence on extensive synthetic data highlights a contrast with PNN6, which delivers superior performance using smaller datasets and fewer resources.

\par As outlined in Table~\ref{tab:OtherTech}, PNN6 achieves comparable or superior results relative to advanced techniques such as UFADPC \cite{c20}, CORAL \cite{c9}, TCNN \cite{c8}, UDTL \cite{c2}, and AMCMENet \cite{c55}. CORrelation ALignment (CORAL) \cite{c9}, a CNN-based deep model, uses transfer learning to minimize the marginal distribution discrepancy between source and target domains. While CORAL, a nine-layer model, performs comparably on the CWRU and JNU datasets, its performance on the PB and SEU datasets is lower than that of PNN6.

\par Notably, the NCVAE-AFL model (Normalized Conditional Variational Auto-Encoder with Adaptive Focal Loss) \cite{c52}, designed for class-imbalanced datasets, achieves high accuracy at training-testing ratios of $20\%$ and $50\%$. However, its large encoder-decoder VAE-based architecture necessitates significantly more training epochs compared to PNN6. Similarly, AMCMENet (Adaptive Multiscale Convolution Manifold Embedding Networks) \cite{c55}, which integrates a 12-layer DNN with LSDA (Locality Sensitive Discriminant Analysis) \cite{c57} for secondary feature extraction and particle swarm optimization for parameter tuning, achieves competitive performance. Nevertheless, AMCMENet's large model size demands substantial training data and computational resources. In contrast, PNN6’s six-layer compact architecture delivers comparable or better accuracy with substantially fewer parameters and significantly reduced training time.

\par Most state-of-the-art models rely on larger architectures that require extensive computational resources and significant data for effective training. In comparison, PNN6 demonstrates robust performance across diverse datasets, including CWRU, PB, JNU, and SEU, while maintaining a smaller model size, faster training times, and reduced computational overhead.

\setlength{\tabcolsep}{6pt} 
\renewcommand{\arraystretch}{1} 
\begin{table}[h!]
\centering
\caption{Performance comparison of proposed PNN6 with latest DNN-based techniques from the literature. The training and testing division ratios (DR) are represented as $10-90\%$ ($10$), $25-75\%$ ($25$), $75-25\%$ ($75$), $80-20\%$ ($80$), $50-50\%$ ($50$) with accuracy in \%.}
 \label{tab:OtherTech}
\renewcommand{\arraystretch}{1.1} 
  \resizebox{\columnwidth}{!}{%
\begin{tabular}{ | c | c | c | c |c|}
\hline
\begin{tabular}[c]{@{}l@{}}Techniques \end {tabular} & Dataset & DR & \begin{tabular}[c]{@{}l@{}} Accuracy \end {tabular}  &   \begin{tabular}[c]{@{}l@{}} PNN6 \\ Accuracy  \end {tabular}\\
\hline
FDFSL \cite{c7}& CWRU &  FSL  &  82.80 &   $100$ \\
\hline

TCNN \cite{c8} &CWRU &  $75$ & $99.9$ & $100$ \\
\hline
 UDTL \cite{c2}& \begin{tabular}[c]{@{}l@{}} CWRU, PB,\\ JNU, \& SEU \end{tabular}& 80 & \begin{tabular}[c]{@{}l@{}}$99.93$, $59.29$\\$97.73$, $57.68$   \end{tabular}  & \begin{tabular}[c]{@{}l@{}}$100$, $98.6$\\$100$, $95.2$   \end{tabular} \\
\hline
CORAL \cite{c9} & CWRU, JNU & $75$ & $97.85$   &$100$,$100$  \\
\hline
UFADPC \cite{c20} & CWRU & $50$ & $85$ &   $100$   \\
\hline
NCVAE-AFL \cite{c52}&CWRU &  $20, 50$ & $96, 100$  & $100, 100$ \\
\hline
\multirow{2}{*}{TabPFN \cite{c48}} & CWRU & $10$ & $89.95$  & $100$ \\
 \cline{2-5}
  & CWRU & $25$ & $93.61$  & $100$ \\

\hline
AMCMENet \cite{c55} &CWRU &  $80$ & $100$ & $100$ \\
\hline

\end{tabular}
}
\end{table}

\subsection {Performance evaluation of PNN in the practical scenario}\label{sec:TL}

\par In industrial settings, data scarcity is a common challenge, hindering the development of highly accurate machine learning models. Transfer Learning (TL) offers a valuable solution by leveraging knowledge acquired from related datasets to enhance performance on target domains with limited data. This study investigates the effectiveness of PNN-based models within a cross-domain TL framework, where the source and target datasets exhibit distinct characteristics, and the target data is extremely limited.

\par Specifically, we evaluate the performance of PNN6, as the fully connected (FC) layer of pre-trained convolutional neural networks (CNNs), namely ResNet18 (RPNN) and VGG11 (VPNN). In our experiments, models trained on a large-scale source dataset (ICE) were fine-tuned on significantly smaller target datasets (JNU and SEU) with limited training data (training-to-testing ratio of 10-90\%).

 \setlength{\tabcolsep}{6pt} 
\renewcommand{\arraystretch}{1} 
\begin{table}[]
\centering
\tiny
\caption{Effect of cross-domain transfer learning in classification accuracy for $10-90\%$ training-testing division ratio in a typical run. The content of each cell is [dataset, accuracy before/accuracy after transfer learning].} 
\label{tab:TL}
\renewcommand{\arraystretch}{1.2} 
\begin{tabular}{ | c | c |c|}
\hline
Target Dataset &  RPNN & VPNN\\
\hline

 JNU &  \begin{tabular}[c]{@{}l@{}}{ICE,69.9,80.6}\end{tabular} & \begin{tabular}[c]{@{}l@{}}{Sim,38.4,95.8}\end{tabular}\\
\hline
 MFPT &  \begin{tabular}[c]{@{}l@{}}{UoC,70.4,79.6}\end{tabular} & \begin{tabular}[c]{@{}l@{}}{SEU,30.9,100}\end{tabular}\\
\hline
\end{tabular}
\end{table}
\par Results demonstrate a substantial performance gain through TL. For instance, on the JNU dataset, accuracy improved from 69.9\% to 80.6\% when using RPNN with TL. Similarly,  a significant accuracy boost on the SEU dataset, from 91.2\% to 96.3\%. With VPNN model, the accuracy on MFPT dataset increased from 30.9\% to 100\% by use of TL on SEU trained model. These findings underscore the effectiveness of PNN-based models for TL in data-constrained industrial applications.

\par The success of this approach can be attributed to the synergy between the robust feature extraction capabilities of pre-trained CNNs and the ability of PNN6 to adapt effectively to new data with minimal fine-tuning. This highlights the potential of PNN-based models as a promising solution for addressing data scarcity challenges in various industrial domains.

\section{PNN architecture Capability Analysis }\label{sec:ArchitectureCapability}
In this section, we perform a comprehensive analysis of the PNN architecture to evaluate its performance on the \emph{ICE} dataset. The analysis encompasses key aspects such as accuracy, loss behavior, gradient dynamics, convergence rate, and feature learning capabilities. These evaluations aim to provide a detailed understanding of the strengths and effectiveness of the PNN-based approach in addressing challenges posed by this dataset.

\subsection{Learning Capabilities of PNN}\label{sec:LearningCapabilities}

The learning capabilities of the proposed PNN6 architecture were analyzed by examining its loss, accuracy, and gradient values during training. Figure \ref{fig:ItrAccLossGradPNN6Tg25} illustrates the evolution of these metrics across training epochs. The results demonstrate that the loss and gradient values converge rapidly to near-zero as training progresses while accuracy steadily approaches $100\%$. 

These observations indicate that the PNN6 architecture has strong learning capabilities and is effective in avoiding the vanishing gradient problem. For visualization, all parameter values within each epoch were normalized.

\begin{figure}[!ht]

\begin{tikzpicture}[tight background]
\begin{axis}[
width=3.5in,
    height=2.250in,
     ylabel = {[Accuracy, Loss, Gradient]},
     xlabel=Iterations,
     xmin=1,
     xmax=185,
     xtick={1,40,80,120,160,185},
     xticklabels={1,40,80,120,160,185},
     x tick label style={font=\tiny},
     y tick label style={font=\tiny},
     ymajorgrids=true,xmajorgrids=true,
grid style=dotted,legend columns=4,
legend style={font=\footnotesize,  at={(0.62, 0.10)}, anchor=south},
scale=1.0
]

\pgfplotstableread{AccNormalized1Col.dat}{\pistonkinetics}
\addlegendentry{Accuracy}  
\addplot [green,mark=., line width=0.75pt] table [x={Iter}, y={Acc}] {\pistonkinetics};

\addlegendentry{Loss}  
\pgfplotstableread{LossNormalized1Col.dat}{\pistonkinetics}
\addplot [red,mark=squre, line width=0.75pt] table [x={Iter}, y={Loss}] {\pistonkinetics};
\addlegendentry{Gradient}  
\pgfplotstableread{GradNormalized1Col.dat}{\pistonkinetics}
\addplot [blue,mark=., line width=0.75pt] table [x={Iter}, y={Gradient}] {\pistonkinetics};
\end{axis}
\end{tikzpicture}
\caption{PNN6 performance metric with epoch ($37$ iterations per epoch) for $75-25\%$ training-testing division ratio.}
\label{fig:ItrAccLossGradPNN6Tg25} 
\end{figure}
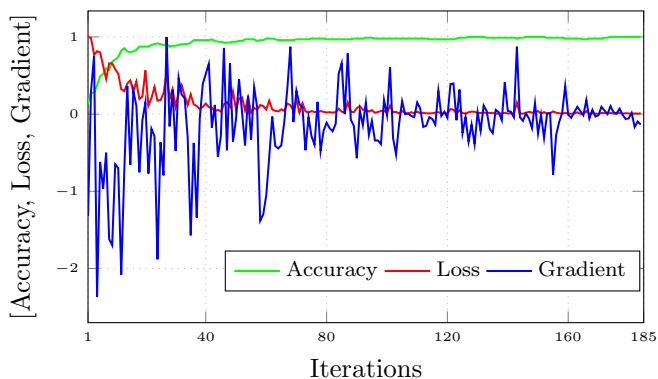

\begin{figure}[!ht]
    \centering
\begin{tikzpicture}          

\begin{axis}[
width=2.9in,
    height=2.25in,
    xlabel =  Epochs,
    ylabel={[Accuracy, Loss, Gradient]},
    xmin=1, xmax=9,
    ymin=-1.2, ymax=1.2,
    xtick={1,2,3,4,5,6,7,8,9},
    ytick={-1.2,-1.0, -0.8, -0.6, -0.4, -0.2, 0.0, 0.2, 0.4, 0.6, 0.8, 1.0, 1.2},  
     x tick label style={font=\tiny},
     y tick label style={font=\tiny},
     ymajorgrids=true,xmajorgrids=true,
grid style=dashed,legend columns=4,
legend style={font=\scriptsize,at={(0.63,0.32)},anchor=north},scale=1.30
  ]
 
\addlegendimage{empty legend}
\addlegendentry{PNN6:}

            \addplot[black,
            mark=triangle,line width=0.75pt]
            coordinates{(1,-1.0)	(2,-0.05)	(3,0)	(4,0.02)};
            \addlegendentry{Grad}
            
            \addplot[red,
            mark=triangle,line width=0.75pt]
            coordinates{(1,1)	(2,0.26)	(3,0.37)	(4,0.23)};
            \addlegendentry{Loss}
            
            \addplot[blue,
            mark=triangle, line width=0.75pt]
            coordinates{(1,0.98)	(2,1)	(3,0.99)	(4,1)};
            \addlegendentry{Acc} 
          \addlegendimage{empty legend}
    \addlegendentry{VDNN6:}
            \addplot[black,
            ultra thick,dotted,line width=0.75pt]
            coordinates{(1,-0.96345038)  (2,0.10547611) (3,-0.04000802) (4,-0.00421713)  (5,0.19179116) (6,-0.10169553) (7, 0.0800789)  (8, 0.03552967) (9, 0.06495316)};
            \addlegendentry{Grad}        
            \addplot[red,
            ultra thick,dotted,line width=0.75pt] 
            coordinates{(1,0.8061) (2,0.8737) (3,0.8690) (4,0.8708) (5,0.8027) (6,0.8697) (7,0.9358) (8,0.8670) (9,1.0000)};
            \addlegendentry{Loss}            
            \addplot[blue,
            ultra thick,dotted, line width=0.75pt]
            coordinates{(1,0.2069) (2,0.0690) (3,0.2069) (4,0.0000) (5,0.0690) (6,0.3103) (7,0.2414) (8,0.1724) (9,0.0690)};
            \addlegendentry{Acc}  
            
\addlegendimage{empty legend}
\addlegendentry{PNN3:}
            
            \addplot[black,
            mark=o, line width=0.75pt]
            coordinates{(1,-.23)	(2,-0.22)	(3,-0.78)	(4,-0.27)	(5,0.14)	(6,-0.25)	(7,-0.17)	(8,0.13)	(9,0.16)};
            \addlegendentry{Grad}
            
            \addplot[red,
            mark=o, line width=0.75pt]
            coordinates{(1,1)	(2,0.48)	(3,0.46)	(4,0.23)	(5,0.26)	(6,0.13)	(7,0.09)	(8,0.08)	(9,0.08)};
            \addlegendentry{Loss}
            
            \addplot[blue,
            mark=o, line width=0.75pt]
            coordinates{(1,0.47)	(2,0.64)	(3,0.75)	(4,0.87)	(5,0.98)	(6,0.99)	(7,1)	(8,0.99)	(9,1)};
            \addlegendentry{Acc}

             \addlegendimage{empty legend}
            \addlegendentry{VDNN3:}
            \addplot[black,
            ultra thick,dashed,line width=0.75pt]
             coordinates{(1,-0.92586654) (2,-0.06096496) (3,-0.00989283)  (4,0.08553732) (5,-0.06709173) (6,-0.08876059) (7,-0.25008646) (8, 0.11187053) (9,-0.21024256)};
            \addlegendentry{Grad}            
            \addplot[red,
            ultra thick, dashed,line width=0.75pt] 
            coordinates{(1,1.0000) (2,0.8641) (3,0.9359) (4,0.8025) (5,0.7960) (6,0.8612) (7,0.8634) (8,0.8638) (9,0.9316)};
            \addlegendentry{Loss}       
            \addplot[blue,
            ultra thick,dashed, line width=0.75pt]
            coordinates{(1,0.2903) (2,0.2258) (3,0.0968) (4,0.1935) (5,0.1935) (6,0.3871) (7,0.1613) (8,0.0000) (9,0.4194)};
            \addlegendentry{Acc}  
            
        \end{axis}

    \end{tikzpicture}
\caption{Accuracy (Acc), loss, and gradients (Grad) performance comparison of the [PNN3 vs VDNN3] \&  [PNN6 vs VDNN6]. The training-testing data division ratio was 75-25\%. The PNN-based models [PNN3 and PNN6] show superior performance compared to VDNN6 and VDNN3 in all three parameters with small training epochs.}
\label{fig:AccLossGradPNN6PNN3ANN6ANN3}
 \end{figure}
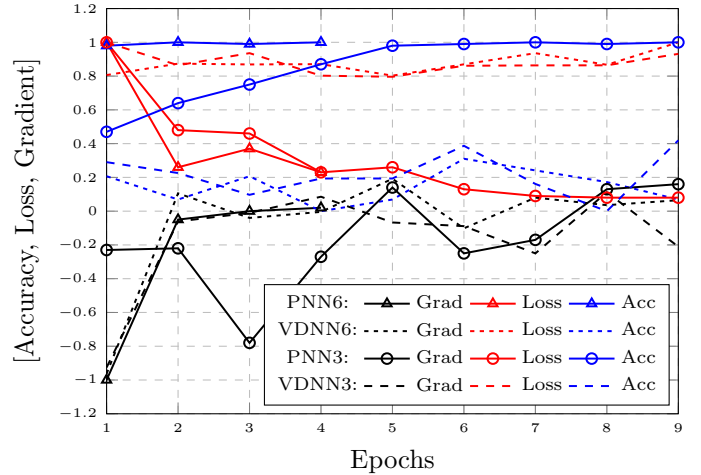

\subsection{Effect of Depth on Learning Capabilities of PNN}\label{sec:EffectOfDepth}

The effect of model depth, i.e., the number of layers, on the learning capabilities of the PNN was analyzed. Figure \ref{fig:AccLossGradPNN6PNN3ANN6ANN3} presents the accuracy, loss, and gradient plots for PNN architectures with six layers (PNN6) and three layers (PNN3), as well as VDNN architectures with six layers (VDNN6) and three layers (VDNN3), for a $75\%-25\%$ training-to-testing ratio. 

In these plots, the gradients are summed across all layers for each epoch and normalized for comparison. The results show that PNN6 achieves a maximum accuracy of $100\%$ within just four epochs. In contrast, PNN3 experiences larger loss and gradient values during the initial epochs, requiring more iterations to reach the same accuracy level. 

For VDNN6 and VDNN3, the loss values remain consistently high throughout the training process. As a result, the gradients do not stabilize, even after nine epochs, leading to lower accuracy. These observations indicate that deeper PNN models, such as PNN6, not only achieve higher accuracy but also require fewer epochs for convergence. This underscores the superior feature-learning capability of deeper PNN architectures compared to shallower ones.

\subsection{Convergence in PNN}\label{sec:Convergence}
\begin{figure}[h!]
  \centering
   \includegraphics[width=\columnwidth]{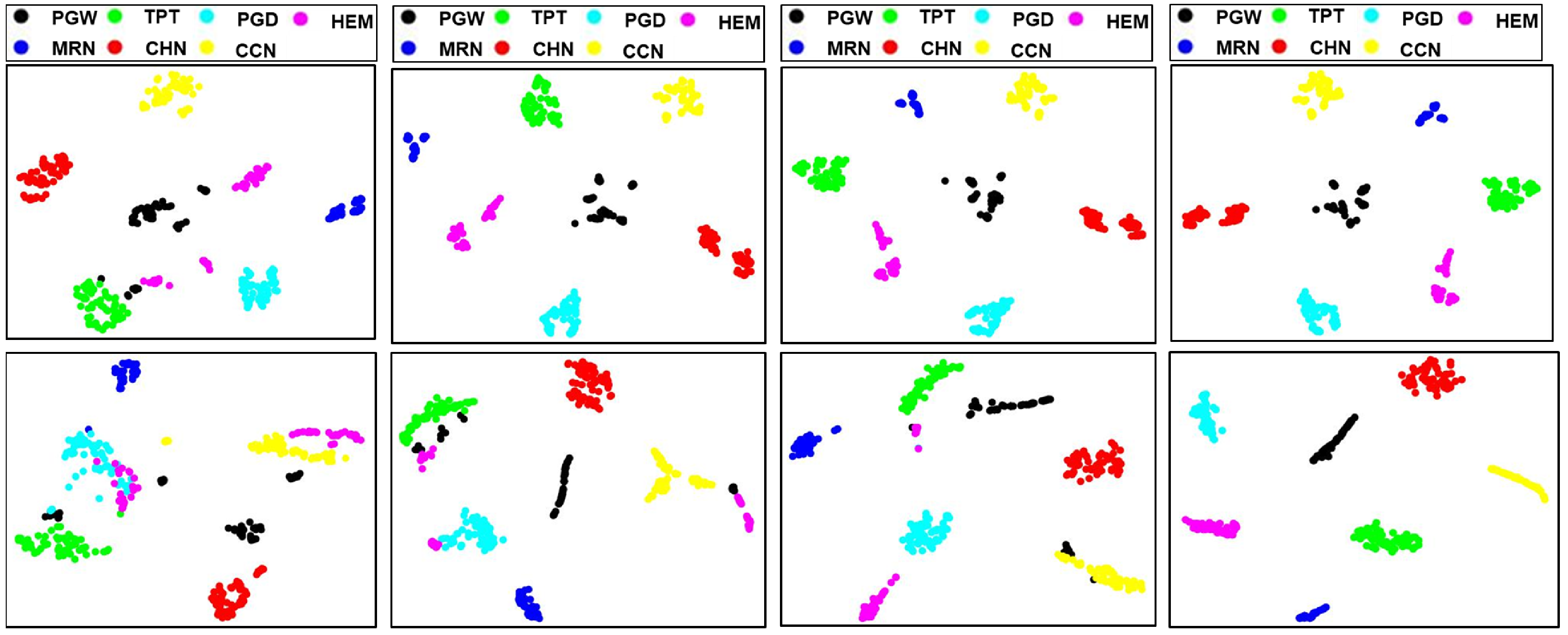}
   \centerline{\footnotesize{ Epoch $1$ \hspace{0.95cm}  Epoch $5$ \hspace{0.99cm} Epoch $10$\hspace{0.95cm}  Epoch $15$}}
  \caption{Comparison of feature learning capability of PNN6 and VDNN6 (in terms of t-SNE plots) for ICE dataset for different iterations of $75-25\%$ training-testing division ratio. (Row 1 for the PNN6 model with Fault ID's as per Table \ref{tab:enginedataset}, and row 2 for the VDNN6 model)}
  \label{fig:EngineEpochsTSNETest25}
\end{figure}

To assess the convergence capabilities of the PNN, we present t-SNE visualizations of the fault features at different training iterations. Figure \ref{fig:EngineEpochsTSNETest25} shows the t-SNE plots of features learned during training by the $5^{\text{th}}$ layer of both PNN and VDNN for different epochs, using the classic $75\%-25\%$ training-to-testing division ratio.

The results demonstrate that the PNN exhibits clear and distinct clustering of fault features within the first five epochs, with the clustering stabilizing thereafter. In contrast, the VDNN model struggles to form well-defined clusters, even after several epochs, which indicates poor feature learning and slower convergence. This comparison highlights the superior convergence behavior and feature learning capabilities of the PNN model.

\begin{figure}[h!]
  \centering
   \includegraphics[width=\columnwidth]{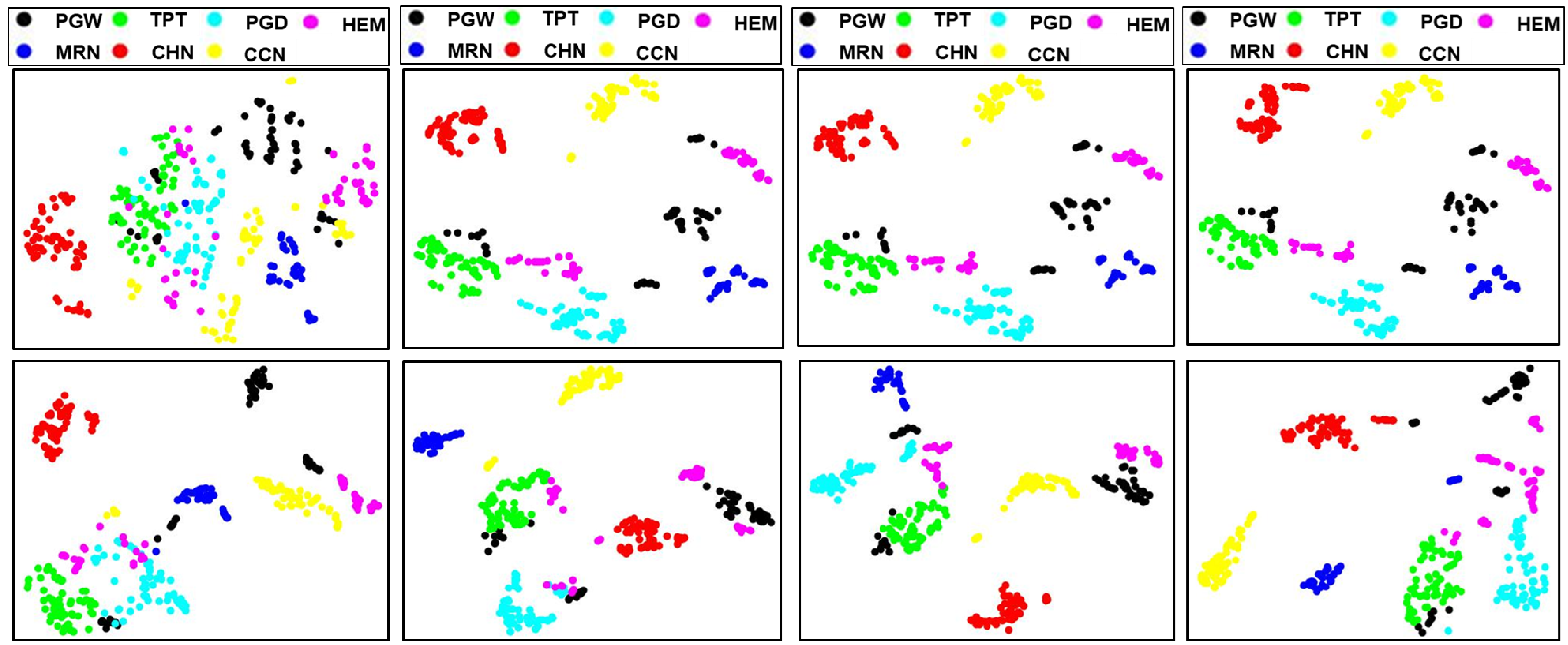}
   \centerline{\footnotesize{ Epoch $1$ \hspace{0.95cm}  Epoch $10$ \hspace{0.99cm} Epoch $20$ \hspace{0.95cm}  Epoch $30$}}
  \caption{Comparison of feature learning capability of PNN6 and VDNN6 for Engine dataset for different iterations of $10-90\%$ training-testing division ratio. (Row 1 for the PNN6 model with Fault ID's as per Table \ref{tab:enginedataset}, and row 2 for the VDNN6 model) }
  \label{fig:EngineEpochsTSNETest90}
\end{figure}

Similarly, for the $10\%-90\%$ division ratio, as shown in Figure \ref{fig:EngineEpochsTSNETest90}, the PNN-based model trains effectively, and within ten epochs, the clustering becomes stable, except for fault classes CCN and PGD. In contrast, the VDNN fails to form proper feature clusters even after ten epochs, resulting in poor clustering and lower classification accuracy. The PNN achieves high classification accuracy (greater than $95\%$) for the ICE dataset within just ten epochs due to its efficient feature learning. The PNN's ability to converge quickly and learn better features can be attributed to its small number of trainable parameters and the absence of the vanishing gradient problem. Consequently, the PNN converges faster and outperforms the VDNN in terms of convergence speed and accuracy for both the small $10\%-90\%$ division ratio and the classical $75\%-25\%$ ratio.

\begin{figure*}[h!]
\centerline{
\includegraphics[width=4.50cm,height=3.75cm]{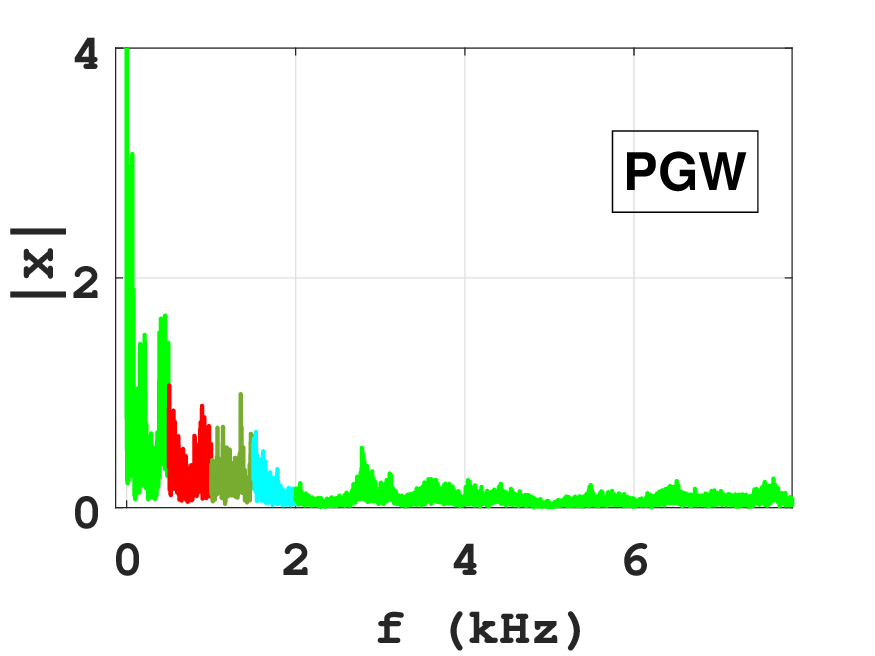}
\includegraphics[width=4.50cm,height=3.75cm]{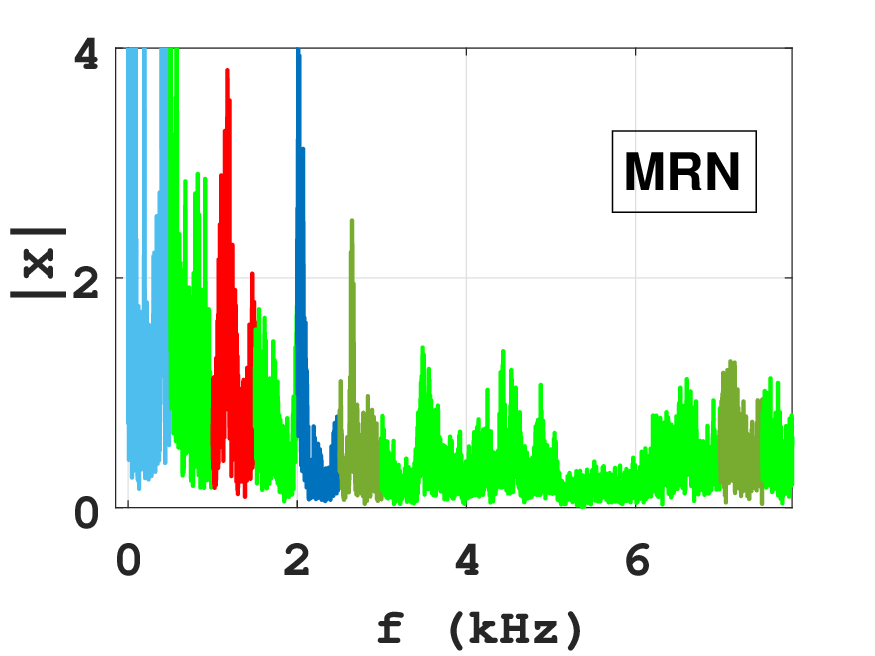}
\includegraphics[width=4.50cm,height=3.75cm]{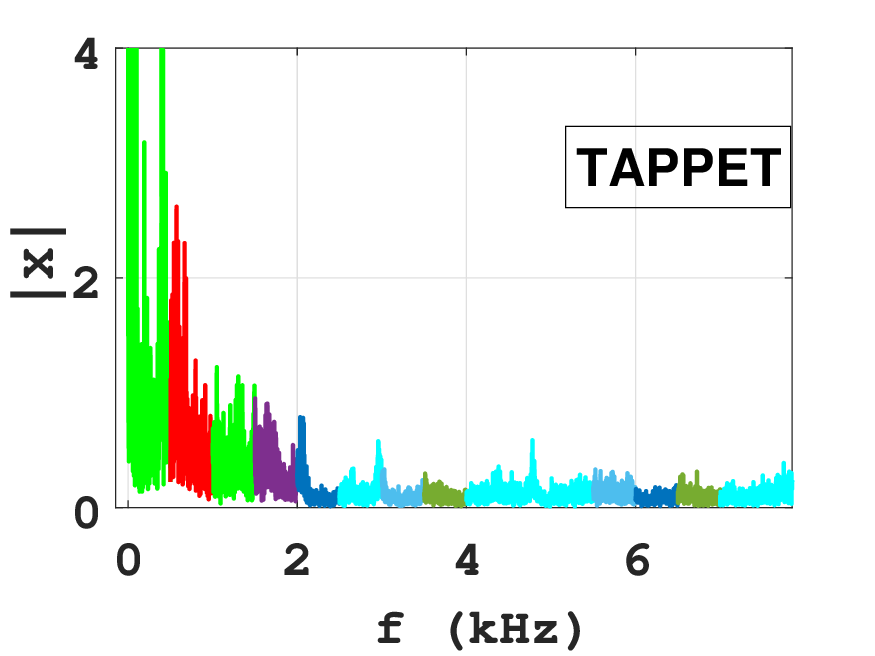}
\includegraphics[width=4.50cm,height=3.75cm]{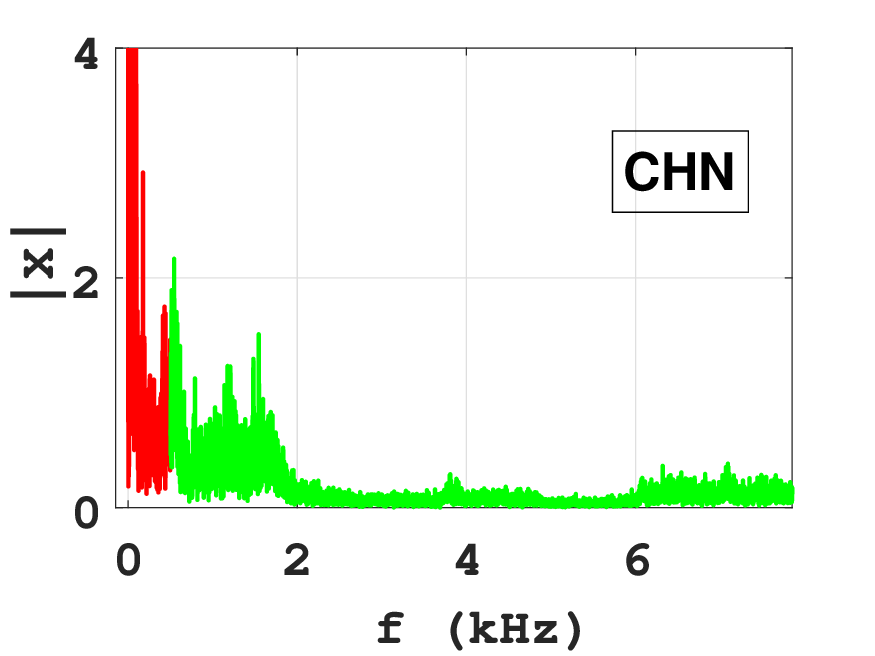}
}
\centerline{
\includegraphics[width=4.50cm,height=3.75cm]{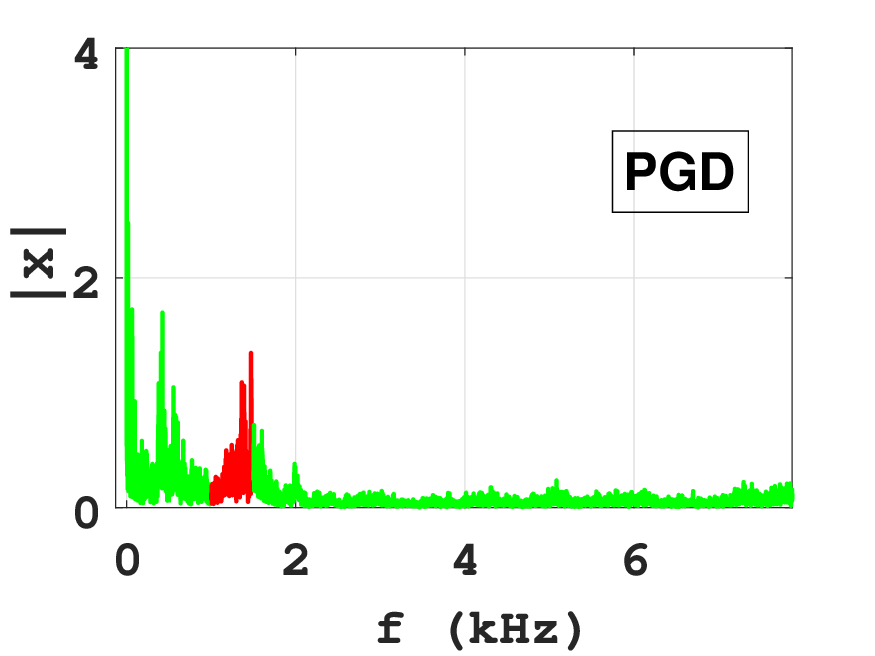}
\includegraphics[width=4.50cm,height=3.75cm]{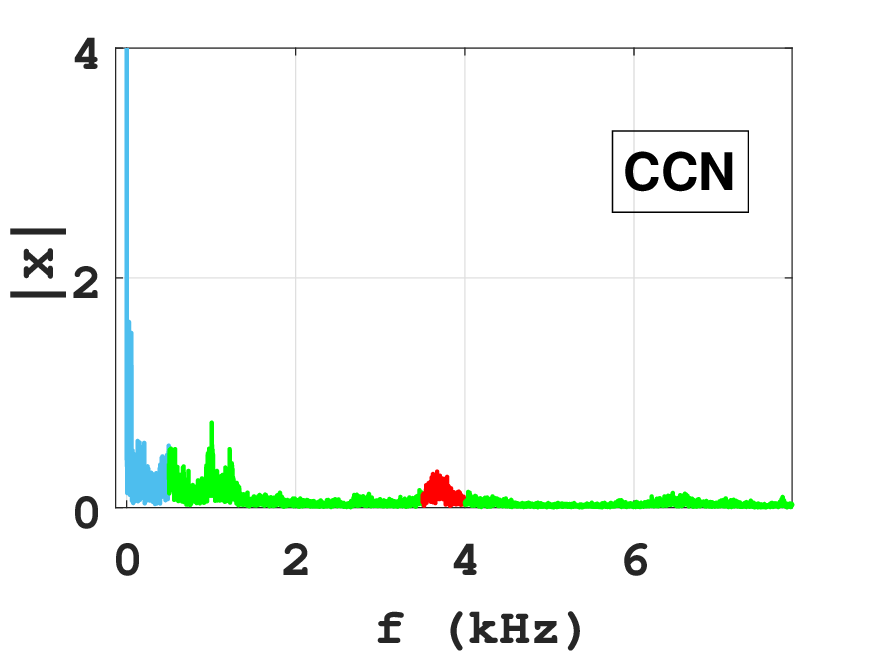}
\includegraphics[width=4.50cm,height=3.75cm]{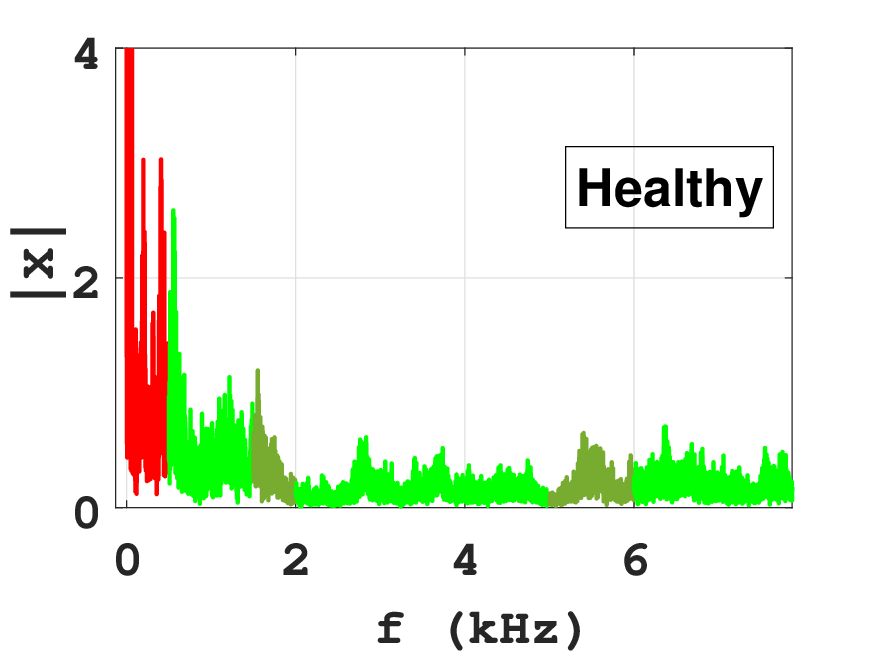}
\includegraphics[width=2cm,height=3.5cm]{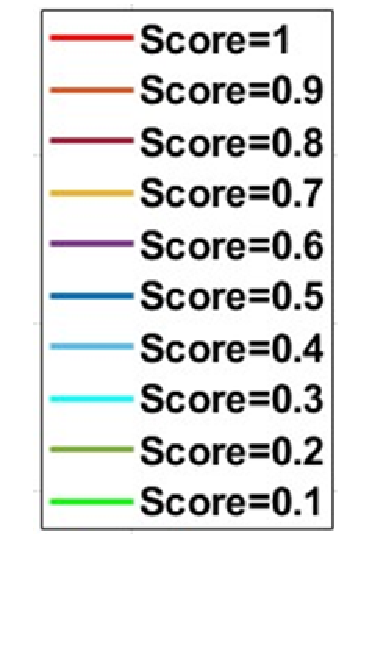}
}

 \caption{ Impact of masking different spectral regions on the softmax score for all fault classes. The spectrum is divided into bins of varying sizes, and the softmax score is recalculated after masking each bin. Red regions indicate spectral areas most critical for fault classification, highlighting the PNN's ability to learn distinct features from specific frequency ranges. These plots are for all faults of the ICE dataset for the mask of size 500 with the training-testing ratio of 75-25\%.}
\label{fig:Mask500SpectrumAllFault}
\end{figure*}

\subsection {Analysis of fault feature and classification performance}\label{sec:FaultFeatureClassification}

The Progressive Neural Network (PNN) effectively learns distinct features for each fault class by leveraging specific sections of the input spectrum. The presence or absence of particular spectral regions significantly impacts the softmax score (classification probability). To analyze this behavior, we applied masking to different portions of the input spectrum using a mask size of $500$ and evaluated the corresponding changes in the softmax score.

\begin{figure*}[h!]
\centerline{
\includegraphics[width=4.2cm, height=3.45cm]{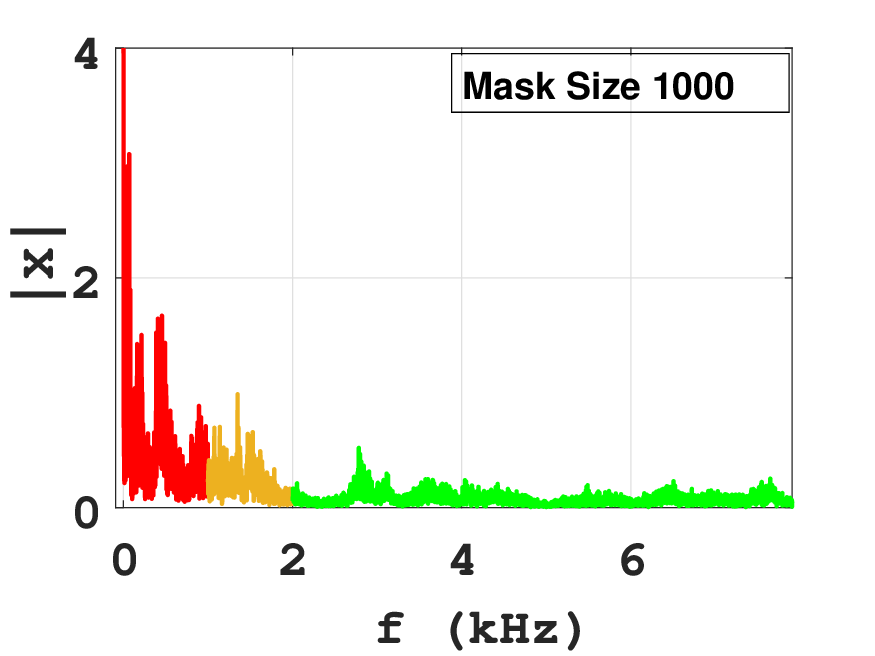}
\includegraphics[width=4.2cm, height=3.45cm]{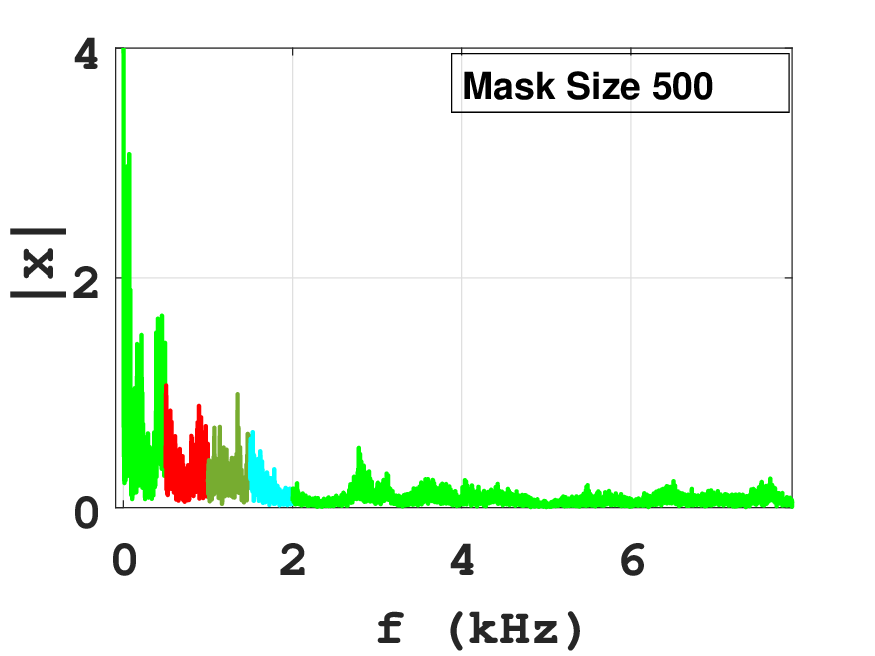}
\includegraphics[width=4.2cm, height=3.45cm]{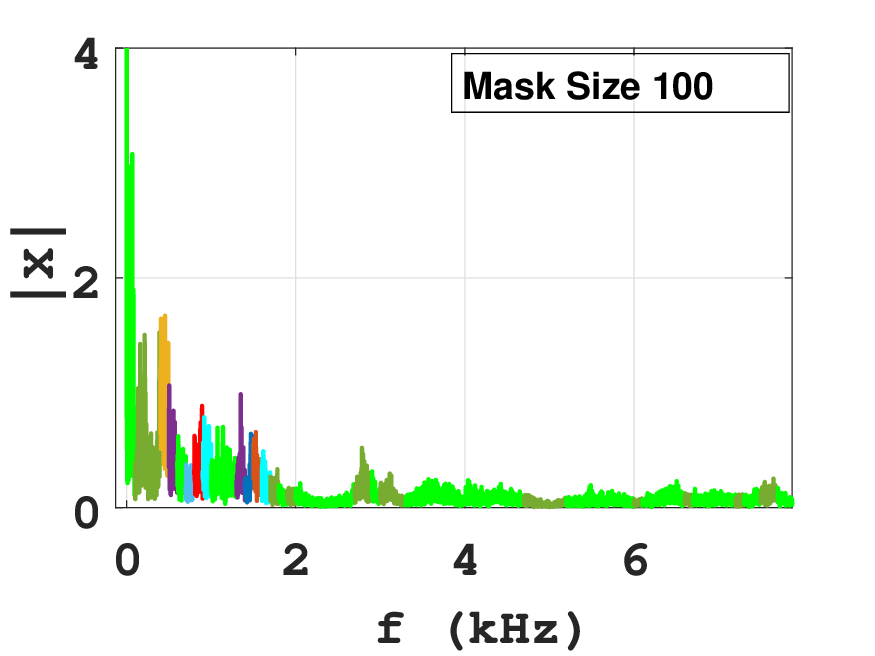}
\includegraphics[width=1.8cm, height=3.3cm]{Legend.eps}
}
\caption{Effect of varying size spectrum masking on the softmax score for the PGW fault class with a training-test ratio of $75$-$25\%$. As the mask size becomes smaller, it reveals finer-grained frequency contributions, highlighting the sensitivity of the PNN to subtle spectral variations.}
\label{fig:Mask}
\end{figure*}

Figure \ref{fig:Mask} illustrates the impact of masking various parts of the spectrum on the softmax score for the PGW fault class. The spectrum was divided into bins of different sizes, and the softmax score was recalculated after masking each bin. Regions marked in red indicate areas of the spectrum responsible for the largest reduction in the softmax score, highlighting their importance in fault classification. This result confirms that the PNN learns distinct features from specific parts of the spectrum for each fault class.

To further refine the analysis, the spectrum was divided into smaller bins, with sizes ranging from $1000$ to $100$. As shown in Figure \ref{fig:Mask}, larger bin sizes identify broader spectral regions contributing to classification, whereas smaller bin sizes reveal finer-grained frequency contributions. For instance, with a bin size of $100$, numerous small frequency components jointly contribute to the classification of the PGW fault class. This observation demonstrates that fault classification relies not on a single frequency but on a collection of interrelated frequencies. 

The interaction between fault features and their associated frequencies, along with their harmonics, underscores the complexity of the underlying relationships. These small frequency bins encapsulate the nuanced features that the PNN leverages to achieve high classification accuracy.

In conclusion, this analysis confirms that the PNN captures unique feature sets from distinct spectral regions for each fault class. The ability to learn from diverse spectral sections is a key factor in the model's superior classification performance.

\subsection{Ablation Studies on PNN}
\label{sec:Ablation}

To evaluate the impact of various design choices in the proposed framework on classification accuracy, we conducted comprehensive ablation studies. These studies examine the effects of data standardization, network depth, hidden layer size ($H_d$), and feed-forward input ($X + z_{h}^{n-2}$) on the \emph{ICE dataset}.

\subsubsection{Impact of Data Standardization and Network Depth}
\label{sec:effect_depth}

The application of N-point FFT-based data standardization negatively impacts the accuracy of PNN6 due to information loss, as shown in Table \ref{tab:effectOfDepth}. This degradation is especially prominent when training data size is limited. Figure \ref{fig:Spectrums} demonstrates the effects of standardization on the FFT spectrum. While data standardization aims to create a unified representation across datasets, the FFT representation enhances features, thereby significantly influencing classification performance.

As shown in Table \ref{tab:effectOfDepth}, increasing the depth of PNN consistently improves accuracy across all training-to-testing division ratios. The optimal performance is observed with PNN6, though the accuracy gain diminishes beyond six layers.

\begin{table}[h!]
\centering
\caption{Effect of Data Standardization and Depth of PNN on Accuracy for ICE dataset.}
\label{tab:effectOfDepth}
\renewcommand{\arraystretch}{1.0}
\resizebox{\columnwidth}{!}{
\begin{tabular}{ |c | c | c |c|c | c | c|}
\hline
Division &\multicolumn{2}{c|}{Data Standardization}   & \multicolumn{4}{c|}{Depth of PNN model}\\
\cline{2-7} 
Ratio & Without & With & PNN3 & PNN4 & PNN5 & PNN6   \\
\hline
$75-25$\%   & $98.56$       & $100$      & $99.01$ & $99.06$ & $99.17$ & $100$   \\
$25-75$\%   & $95.60$       & $96.95$    & $96.45$ & $96.48$ & $96.48$ & $96.55$    \\
$10-90$\%   & $87.78$       & $93.60$    & $92.61$ & $92.44$ & $93.36$ & $93.60$    \\
\hline
\end{tabular}
}
\end{table}

\subsubsection{Impact of Feed-Forward Input and Hidden Layer Output}
\label{sec:effect_feedforward}

\begin{figure}[!ht]
\centering
\includegraphics[width=3in, height=3.1in]{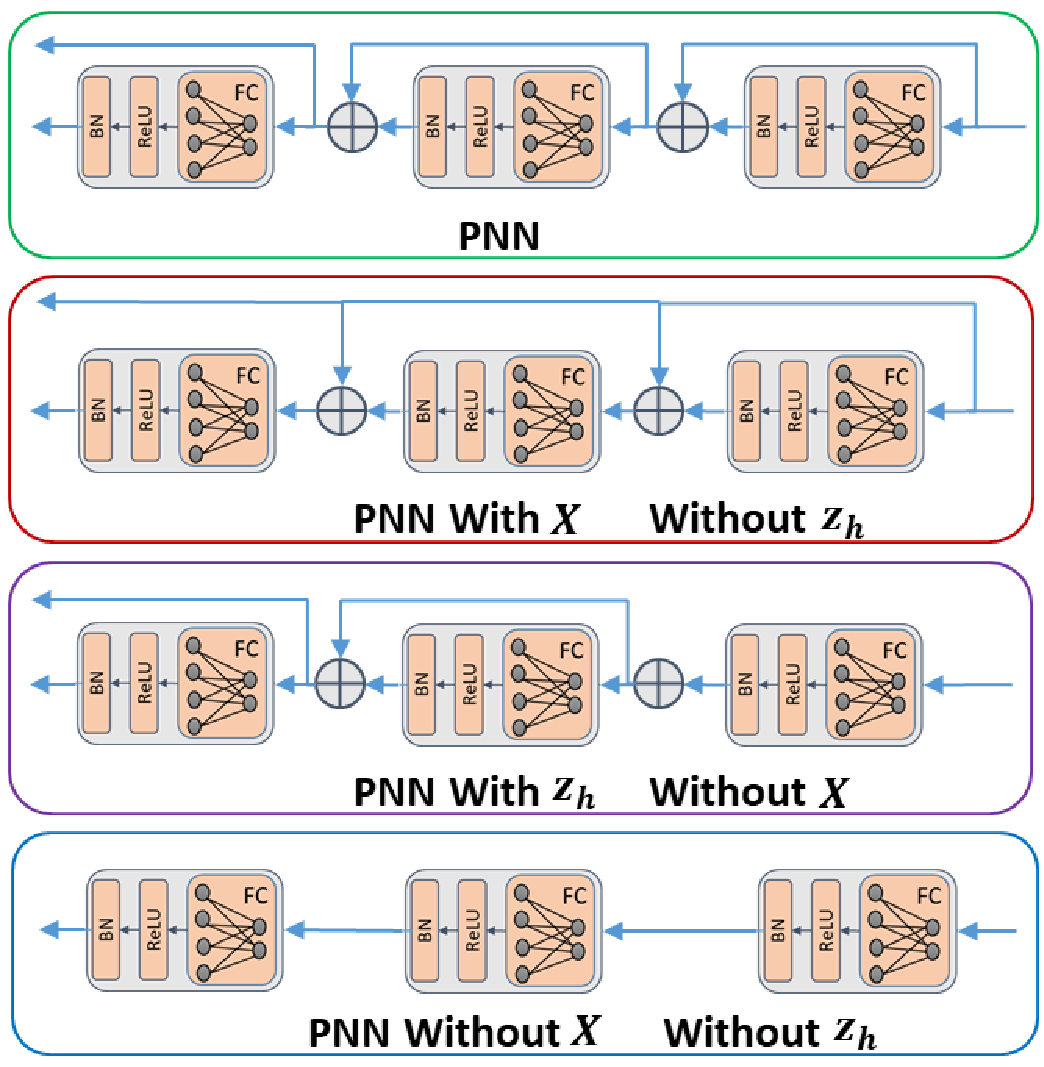}
\caption{PNN Architectures with and without $X$ and $z_{h}$}
\label{fig:ModifiedPNN}
\end{figure}

We analyzed the role of feed-forward input ($X + z_{h}^{n-2}$) with a size of $(K + (n-2)H_d)$ in the $75-25\%$ division ratio, as detailed in Table \ref{tab:DirectInputRemovalEffect}. The impact of removing either or both components ($X$ and $z_{h}$) is illustrated in Figure \ref{fig:ModifiedPNN}.

The removal of both components reduces accuracy to $33.3\%$, effectively transforming the model into a VDNN, as gradients cannot propagate directly from the output layer to earlier layers. When $z_{h}$ is excluded from the feed-forward input, accuracy drops to $91.67\%$ due to reduced capability in estimating higher-order features. Similarly, excluding $X$ alone causes a significant accuracy loss to $15.63\%$, highlighting the complementary roles of both components.

\begin{table}[h]
\centering
\caption{Effect of Feed-Forward Connections ($X + z_{h}^{n-2}$) on Accuracy for $75-25\%$ Division Ratio for ICE dataset}
\label{tab:DirectInputRemovalEffect}
\renewcommand{\arraystretch}{1.0}
\begin{tabular}{ | c | c | c | c |}
\hline
\begin{tabular}[c]{@{}l@{}}With \\ $X + z_{h}^{n-2}$\end{tabular} & \begin{tabular}[c]{@{}l@{}}With $X$, \\ Without $z_{h}^{n-2}$ \end{tabular} & \begin{tabular}[c]{@{}l@{}}With $z_{h}^{n-2}$, \\ Without $X$\end{tabular} & \begin{tabular}[c]{@{}l@{}}Without $X$, \\ Without $z_{h}^{n-2}$ \end{tabular} \\
\hline
$100$  & $91.67$ &  $15.63$ & $33.3$  \\
\hline
\end{tabular}
\end{table}

\subsubsection{Effect of Hidden Layer Size}
\label{sec:effect_hidden_size}

We also evaluated the influence of hidden layer size ($H_d$) on PNN6 performance, as summarized in Table \ref{tab:effectOfHiddenSize}. The results show minimal improvement in accuracy beyond $H_d = 100$. Most of the classification accuracy stems from the direct input ($X + z_{h}^{n-2}$), making the hidden layer size less critical beyond a certain point. Additionally, larger hidden layers tend to shorten training time, as training stops once maximum accuracy is achieved.

\begin{table}[h!]
\tiny
\centering
\caption{Effect of Hidden Layer Size ($H_d$) on PNN Performance for Different Division Ratios for ICE dataset}
\label{tab:effectOfHiddenSize}
\renewcommand{\arraystretch}{1.0}
\resizebox{\columnwidth}{!}{%
\begin{tabular}{ |c | c | c |c|c|c |}
\hline
$H_d$  & \multicolumn{3}{c|}{Division Ratio}  & Time (in sec)   & Model  \\
\cline{2-4}
(\#) & $75\%$ & $25\%$ & $10\%$ & (for DR $25\%$) &  Size (MB)\\
\hline
10 & 98.75 &  96.25 & 92.25  &  1.79 & 3.57  \\    
50 & 98.96 & 96.45  & 92.36  & 1.83  & 16.17 \\   
100 & 99.27 & 96.49  & 92.36  & 2.14  & 32.1  \\   
250 & 99.58 & 96.52  & 94.08  & 2.42  & 81.01 \\     
500 & 99.48 & 96.52  & 94.78  & 5.24  & 166.33 \\     
1000 & 99.89 & 96.38  & 93.08  & 4.32  & 351.31 \\  
\hline
\end{tabular}
}
\end{table}

\subsection{Computational Complexity Analysis}
\label{sec:ComputationalComplexity}

After dimensionality reduction, the standard FFT input vector size was fixed at $N = 16,384$ for all datasets. Using this input, the Vanilla Deep Neural Network (VDNN) model required approximately $178$ million parameters, whereas the PNN model of the same depth required only $8$ million parameters for $H_d = 100$. In the PNN architecture, the hidden layer size $H_d$ is kept constant across all layers, with $H_d \ll N_i$. Consequently, the increment in the number of neurons in successive layers is proportional to $H_d$. This design results in a significantly lower parameter count for PNN compared to a VDNN of the same depth, making it computationally efficient.

Furthermore, the small model size, low training data requirements, and reduced computational complexity make the proposed PNN model suitable for embedded AI applications. Its improved cross-domain transfer learning capabilities also render it applicable across a wide range of industrial scenarios.
\section{Conclusion and Future Work}
\label{sec:Conclusion}

This paper presents an optimal fault classification framework for rotating machinery under conditions of limited data availability. The proposed novel Progressive Neural Network (PNN), has been extensively evaluated on six open-source and two in-house datasets. The results demonstrate that the PNN-based models achieve exceptional fault detection performance, exceeding $99\%$ accuracy in the standard $75$-$25\%$ training-testing division ratio. Even under small-data scenarios ($10$-$90\%$ and $25$-$75\%$), the framework achieves accuracies above $85\%$, highlighting its robustness and effectiveness for fault classification in vibration data from various rotating machinery. Empirical analysis reveals that the PNN architecture facilitates the formation of non-overlapping feature clusters in t-SNE feature plots within fewer iterations compared to classical DNNs. Additionally, the PNN model is lightweight, requiring significantly fewer parameters than vanilla DNN-based models, which reduces computational complexity and enhances training efficiency. The current framework has not been tested on non-rotary machinery or for compound fault classification. Our Future research will focus on extending the proposed approach to address these limitations, thereby broadening its applicability to a wider range of fault types and industrial use cases.



\end{document}